\documentclass[sigconf]{acmart}

\usepackage{multirow}
\usepackage{booktabs}
\usepackage{enumitem}
\usepackage{graphicx}
\usepackage{balance}
\newcommand{\stitle}[1]{\par\noindent\textbf{#1}\ }
\AtBeginDocument{%
  }

\copyrightyear{2026}
\acmYear{2026}
\setcopyright{cc}
\setcctype{by}
\acmConference[MM '26]{Proceedings of the 34th ACM International Conference on Multimedia}{November 10--14, 2026}{Rio de Janeiro, Brazil}
\acmBooktitle{Proceedings of the 34th ACM International Conference on Multimedia (MM '26), November 10--14, 2026, Rio de Janeiro, Brazil}
\acmDOI{10.1145/3767308.3836063}
\acmISBN{979-8-4007-2213-4/2026/11}

\acmSubmissionID{mfp6416}

\begin{document}\sloppy

\title[Learning to Curate What You Generate for Generalizable Few-Shot Class-Incremental Learning]{Learning to Curate What You Generate for Generalizable Few-Shot Class-Incremental Learning}


\author{Junhui Yin}
\authornote{Junhui Yin and Yuchen Yang contributed equally to this work.}
\email{junhuiyin23@gmail.com}
\affiliation{%
  \institution{University of Science and Technology Beijing}%
  \city{Beijing}%
  \country{China}%
}

\author{Yuchen Yang}
\authornotemark[1]
\email{3319733548@bupt.edu.cn}
\affiliation{%
  \institution{Beijing University of Posts and Telecommunications}%
  \city{Beijing}%
  \country{China}%
}

\author{Yilin Yin}
\email{yinyilin2023@bupt.edu.cn}
\affiliation{%
  \institution{Beijing University of Posts and Telecommunications}%
  \city{Beijing}%
  \country{China}%
}

\author{Shuai Na}
\email{nasa65485@bupt.edu.cn}
\affiliation{%
  \institution{Beijing University of Posts and Telecommunications}%
  \city{Beijing}%
  \country{China}%
}

\author{Haoran Xi}
\email{galaxyxhr@bupt.edu.cn}
\affiliation{%
  \institution{Beijing University of Posts and Telecommunications}%
  \city{Beijing}%
  \country{China}%
}

\author{Jianhua Yang}
\authornote{Corresponding author.}
\email{jianhua.yang@cripac.ia.ac.cn}
\affiliation{%
  \institution{Institute of Automation, Chinese Academy of Sciences}%
  \city{Beijing}%
  \country{China}%
}

\author{Muyi Sun}
\email{muyi.sun@bupt.edu.cn}
\affiliation{%
  \institution{Beijing University of Posts and Telecommunications}%
  \city{Beijing}%
  \country{China}%
}

\author{Man Zhang}
\email{zhangman@bupt.edu.cn}
\affiliation{%
  \institution{Beijing University of Posts and Telecommunications}%
  \city{Beijing}%
  \country{China}%
}

\author{Shengfeng He}
\email{shengfenghe7@gmail.com}
\affiliation{%
  \institution{Singapore Management University}%
  \city{Singapore}%
  \country{Singapore}%
}


\renewcommand{\shortauthors}{Junhui Yin et al.}

\begin{abstract}
Few-shot class-incremental learning (FSCIL) aims to learn novel classes from limited annotations while preserving prior knowledge. Existing methods typically assume a sufficiently large base session, but this assumption fails when both base and incremental data are scarce, leading to weak initial representations, semantic drift, and unstable boundaries. We study this underexplored yet realistic setting, termed Generalizable FSCIL (G-FSCIL), where the base session itself contains only a few classes. Although synthetic data can alleviate supervision scarcity, naively mixing generated samples often introduces semantic noise and exacerbates old–new boundary conflicts.
To address this, we propose a framework that curates trustworthy synthetic knowledge for stable G-FSCIL. Specifically, we first construct class-specific synthetic candidate pools using a frozen latent diffusion model, where class inversion is performed at the first observation and the resulting condition embeddings are reused for on-demand generation. Building on these candidates, we learn a knowledge curation strategy that selects samples with both semantic consistency and visual diversity, and distill this process into a transferable selection policy during the base session, which is then reused without further optimization. Leveraging the curated synthetic data, we further design a boundary-stable incremental adaptation scheme, including synthetic-informed prototype initialization and bidirectional boundary calibration to mitigate old–new conflicts. Extensive experiments demonstrate that our method consistently outperforms existing FSCIL baselines, with reduced forgetting and improved balance between old and new classes. Code is available at \url{https://github.com/NiHaoWoJiaoYYC/G-FSCIL}.

\end{abstract}

\ccsdesc[500]{Computing methodologies~Lifelong machine learning}

\keywords{Generalizable FSCIL, synthetic knowledge curation, boundary calibration}


\maketitle

\section{Introduction}\label{sec_intro}

The rapid advancement of artificial intelligence has enabled deep neural networks to achieve remarkable success in visual recognition tasks~\cite{yin2026plug}. However, most models follow a static paradigm with all categories available simultaneously during training. In real-world environments, new categories continually emerge over time, requiring models to incrementally acquire new knowledge while retaining previously learned information. Class-Incremental Learning (CIL)~\cite{rebuffi2017icarl,wang2022beef} addresses this challenge by enabling models to learn new classes without retraining from scratch.
In practice, however, newly introduced classes are often associated with only a handful of labeled samples. This limitation motivates Few-Shot Class-Incremental Learning (FSCIL)~\cite{wang2023few,liu2024few}, which aims to incrementally learn novel classes from extremely limited supervision while preserving previously acquired knowledge. Compared to standard CIL, FSCIL is more vulnerable to catastrophic forgetting~\cite{mccloskey1989catastrophic} and overfitting~\cite{chen2019closer}, due to both scarce data and repeated updates.

\begin{figure}[t]
   \includegraphics[width=\linewidth]{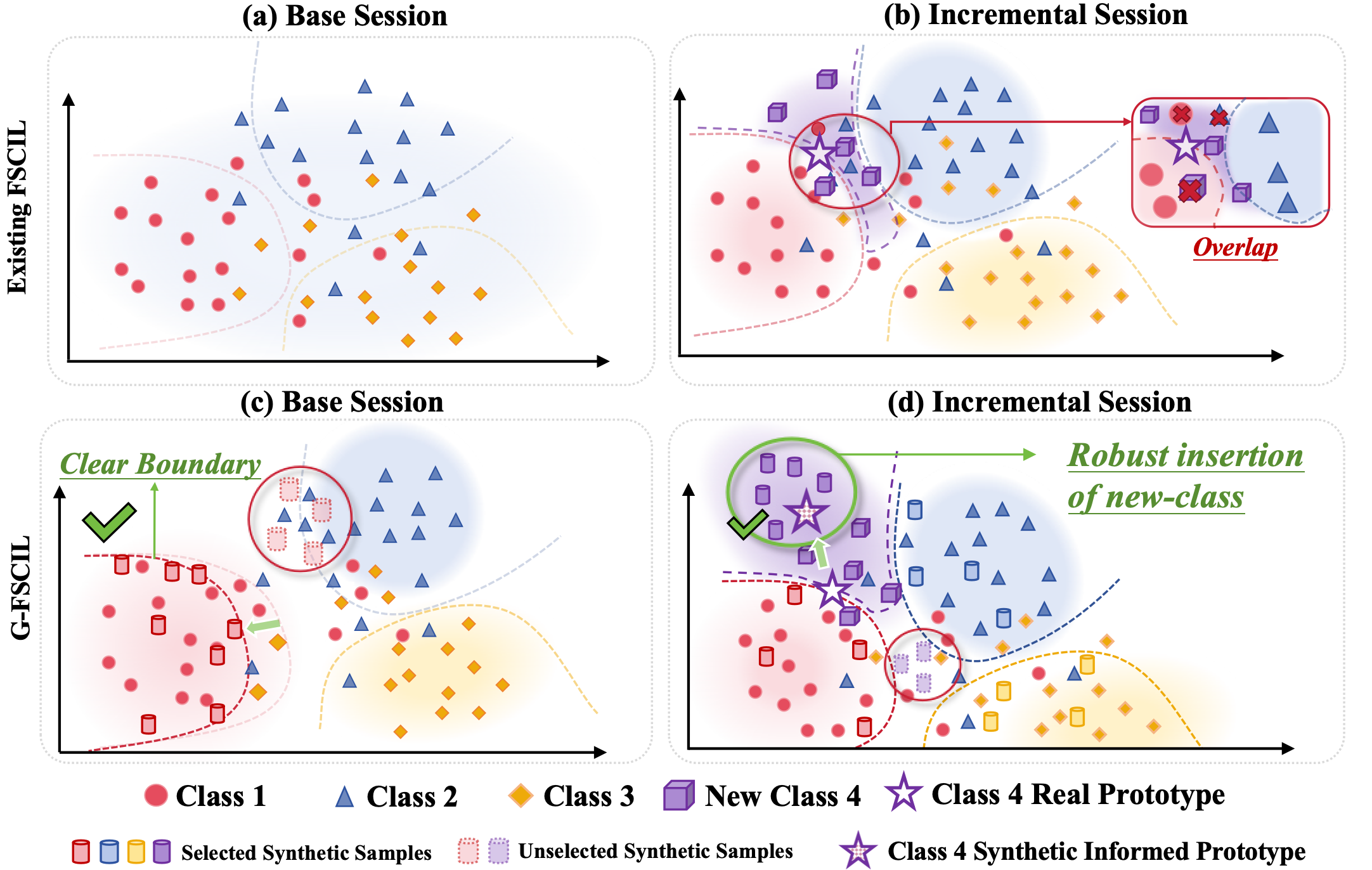}
   \caption{
The comparison of our proposed G-FSCIL with existing FSCIL frameworks.
Top row: Under weak base priors, existing FSCIL methods suffer from scattered feature geometry, unstable novel-class insertion, and severe old--new overlap.
Bottom row: Our method curates trustworthy synthetic supports to refine the feature space and stabilize the insertion of new classes, thereby enabling robust incremental adaptation.}
   \Description{Four-panel comparison of conventional FSCIL and G-FSCIL. Conventional FSCIL has scattered base features and overlap after inserting a novel class. G-FSCIL uses selected synthetic samples to clarify base boundaries and place the novel-class prototype in a separated region.}
   \label{fig:PF}
\end{figure}

To address these challenges, existing FSCIL methods broadly fall into three groups: (i) generative replay methods~\cite{liu2022few, cheraghian2021synthesized, agarwal2022semantics}, which synthesize additional samples to compensate for limited data; (ii) replay-based methods~\cite{li2024crossmodal, zhang2023eprc, ji2023mcnet}, which retain representative historical samples to alleviate forgetting; and (iii) prompt-based methods~\cite{liu2025sec,liu2024few,jiang2025lgsp}, which improve base-session representation learning and facilitate knowledge transfer to incremental stages. Despite their progress, these methods generally rely on a sufficiently large and diverse base session to establish reliable category priors for subsequent adaptation.
However, this assumption is often impractical. For example, privacy constraints may leave only a limited number of users available during face-recognition initialization, yielding a weakly structured base-session feature space. Figure~\ref{fig:PF}(a) shows that class features are scattered and fail to form clear decision boundaries. Consequently, when new classes are introduced, their features can easily overlap with previously occupied regions, leading to severe confusion between old and novel categories. As shown in Figure~\ref{fig:PF}(b), newly inserted class prototypes can significantly distort existing decision boundaries.

To investigate this more realistic scenario, we consider a challenging setting termed \emph{Generalizable FSCIL (G-FSCIL)}, where the base session itself contains only a limited number of classes. A natural way to alleviate supervision scarcity is to leverage synthetic data~\cite{liu2025diffgre,feng2023diverse}, which can enrich intra-class variation and improve prototype estimation. However, naively incorporating synthetic samples often introduces substantial noise. Low-quality generations may cause semantic drift, further disrupt the already fragile feature space, and exacerbate old--new boundary conflicts. This leads to a key question: \emph{how can we identify and exploit trustworthy synthetic knowledge to support stable incremental learning under the G-FSCIL setting?}

To answer this question, we propose a framework that learns to \emph{curate trustworthy synthetic knowledge} for stable G-FSCIL. Specifically, we construct class-specific synthetic candidate pools using a frozen pre-trained latent diffusion model. Instead of training a generator, we perform class inversion only once when each class is first observed, obtaining a compact class-specific condition embedding that is reused in subsequent sessions to synthesize candidate samples on demand.
 During the base session, we learn a curation strategy to select semantically aligned, visually diverse synthetic supports and distill this behavior into a transferable policy directly reusable in subsequent sessions without selector re-optimization under limited supervision.
Based on the curated synthetic samples, we further develop a \emph{boundary-stable incremental adaptation} strategy. As illustrated in Figure~\ref{fig:PF}(c) and (d), we first perform synthetic-informed prototype initialization to stabilize the insertion of new classes, and then apply bidirectional boundary calibration to reduce interference between old and new categories. In this way, the model effectively leverages trustworthy synthetic knowledge to complement scarce real data while preserving previously learned decision regions. In summary, our contributions are fourfold:
\begin{itemize}[leftmargin=*, noitemsep, topsep=0pt]
    
    \item We introduce \emph{Generalizable FSCIL}, a more realistic setting where the base session contains only a limited number of classes, leading to weak category priors and significantly increased difficulty for subsequent incremental learning.

    \item We propose a framework that learns to \emph{curate trustworthy synthetic knowledge}. It constructs class-specific synthetic candidate pools, learns to select semantically aligned and diverse supports during the base session, and distills this process into a transferable selection policy for later sessions.

    \item We develop a \emph{boundary-stable incremental adaptation} strategy that combines synthetic-informed prototype initialization with bidirectional boundary calibration, enabling stable novel-class insertion and mitigating old--new conflicts.

    \item Extensive experiments show consistent improvements over existing methods with higher accuracy and reduced forgetting.
\end{itemize}

\section{Related Work}

\noindent\textbf{Few-Shot Class-Incremental Learning.}
Few-shot class-incremental learning (FSCIL) studies incremental recognition where new classes are introduced with only a few labeled samples. Existing FSCIL methods mainly fall into three categories: (i) classifier or prototype adaptation for novel-class insertion~\cite{zhang2021few,zhou2022forward,wang2023few}; (ii) replay and semantic regularization to mitigate forgetting and overfitting~\cite{li2024crossmodal,zhang2023eprc,song2023learning}; and (iii) prompt-based transfer with pre-trained vision transformers~\cite{wang2022l2p,wang2022dualprompt,smith2023codaprompt,liu2024few,liu2025sec}, together with training-free calibration using multimodal priors~\cite{chen2025bimc}. Recent studies further improve the base feature space by reserving feature capacity, balancing discriminability and transferability, or exploiting stronger vision--language priors~\cite{ahmed2024orco,oh2024closer,park2024pretrained}.
Despite these advances, most FSCIL methods assume that the base session provides sufficiently rich category priors, which may not hold in practice. For example, privacy constraints in face recognition can limit the number and diversity of base classes available at initialization. This motivates our study of G-FSCIL, where incremental learning starts from a weak base prior.

\noindent\textbf{Diffusion-Based Data Generation.}
Diffusion models have been widely used for data synthesis, augmentation, and replay. Prior work uses them to enrich intra-class variation for prompt tuning~\cite{feng2023diverse}, alleviate few-shot scarcity via synthesized features or semantic replay~\cite{cheraghian2021synthesized,agarwal2022semantics,liu2022few}, and approximate historical data distributions in continual learning~\cite{gao2023ddgr}. DDDR~\cite{liang2024dddr} avoids generator retraining via reusable class inversion, while DiffGRE~\cite{liu2025diffgre} emphasizes synthetic-data refinement. Dataset distillation further shows the importance of informative subset selection over data quantity~\cite{cazenavette2022mtt,zhong2025mct}. These results indicate that the key challenge is not generation itself, but ensuring synthetic-data fidelity and utility. Unlike replay-oriented methods that approximate past distributions, our approach builds reusable class-specific knowledge and selectively exploits it under weak base priors.

\noindent\textbf{Data Selection and Synthetic Knowledge Curation.}
Data selection aims to identify compact yet informative subsets based on importance, coverage, or downstream utility, particularly for synthetic data with varying quality.
Recent studies~\cite{gadre2023datacomp,goyal2024scaling,kim2025respec,wei2025hqclip} have shown that data curation is critical for improving training data quality, where effective filtering and relevance-aware selection can be more beneficial than simply increasing data scale or compute. In FSCIL, synthetic data are useful only when they preserve semantic consistency while introducing complementary diversity, as explored through replay~\cite{zhang2023eprc,agarwal2022semantics}, semantic constraints~\cite{song2023learning}, and sample refinement~\cite{liu2025diffgre}. Different from these methods, we explicitly learn semantic- and diversity-aware curation during the base session and distill it into a transferable policy that is reused in incremental sessions.

\section{Methodology}

\begin{figure*}[t]
    \centering
    \includegraphics[width=0.95\linewidth]{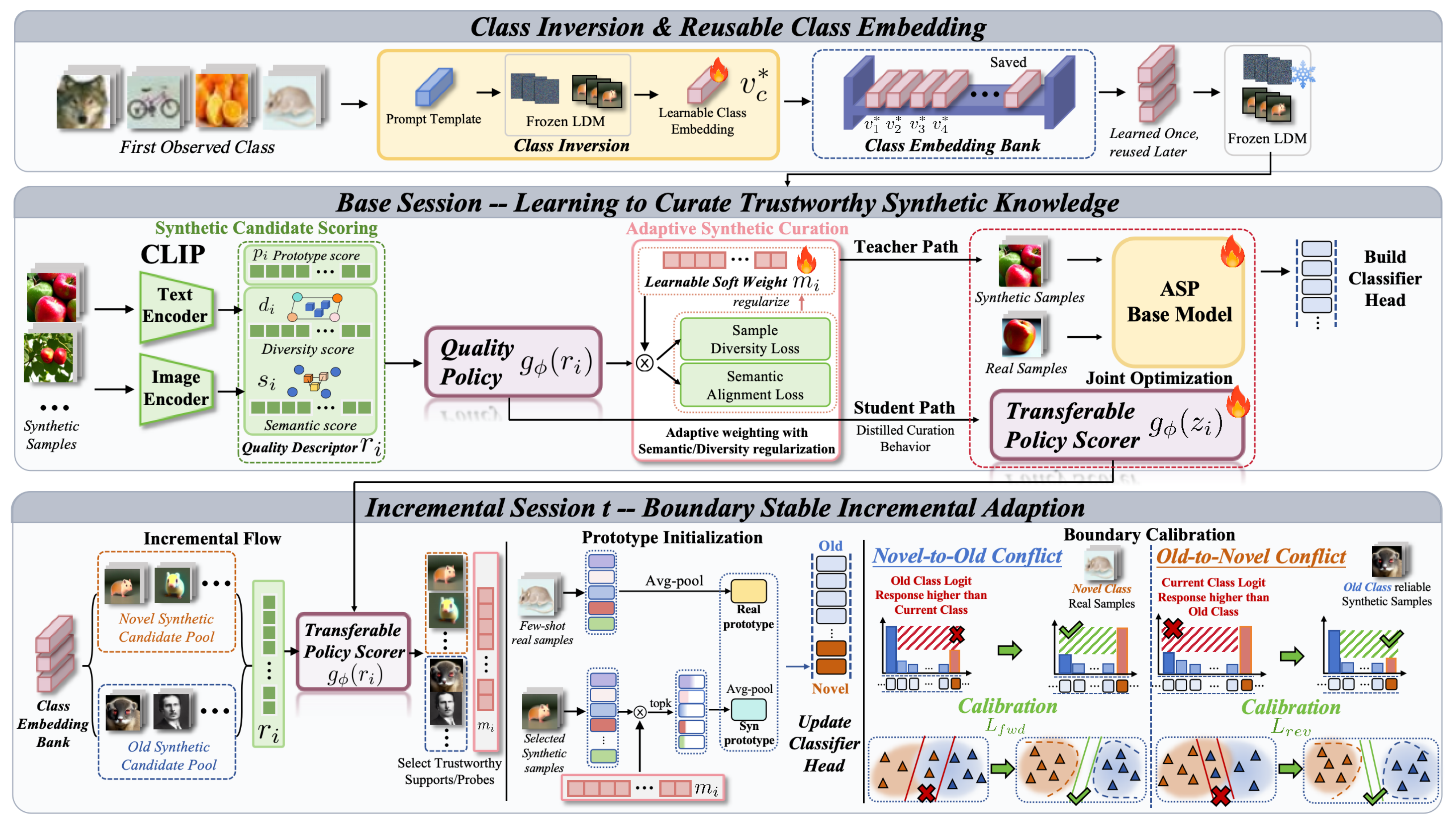}
    \caption{Overview of the proposed G-FSCIL framework. Each class is inverted once upon first observation into a reusable embedding for image synthesis during the current session. During the base session, synthetic candidates
are selectively curated and distilled into a transferable policy. During incremental learning, the stored embeddings generate
candidate pools on demand, and the distilled policy selects reliable synthetic supports to improve prototype initialization and
boundary calibration, thereby enabling more stable novel-class insertion and better old–new balance.
    }

    \Description{Three-stage framework. Stage 1 performs class inversion with a frozen latent diffusion model and stores reusable class embeddings. Stage 2 scores synthetic candidates for semantic consistency and diversity, learns adaptive curation weights, and distills them into a transferable policy scorer. Stage 3 uses the stored embeddings to generate candidates on demand and applies the scorer for synthetic-informed prototype initialization and bidirectional boundary calibration.}
    \label{fig:overview}
\end{figure*} 

{\looseness=-1
\textbf{Problem Setup.}
Unlike conventional FSCIL settings with a relatively large base-class pool, Generalizable FSCIL (G-FSCIL) considers a more challenging scenario in which the base session contains only a small number of classes and the full task is divided into more incremental sessions. Consequently, the model starts from substantially weaker category priors, and each subsequent session must adapt novel classes under a more challenging incremental process.
\par}

Formally, let $\{\mathcal{D}_{\mathrm{train}}^{0}, \mathcal{D}_{\mathrm{train}}^{1}, \ldots, \mathcal{D}_{\mathrm{train}}^{T}\}$ denote the sequence of labeled training sets across $T{+}1$ sessions, where $\mathcal{D}_{\mathrm{train}}^{t}$ is associated with the class set $\mathcal{C}_t$ at session $t$. The class sets are mutually disjoint across sessions, i.e.,
$
\mathcal{C}_i \cap \mathcal{C}_j = \emptyset, \quad \forall i \neq j.
$
In the base session ($t=0$), the model is initialized using $\mathcal{D}_{\mathrm{train}}^{0}$ with a limited base-class set $\mathcal{C}_0$, where $|\mathcal{C}_0| = B$. For each subsequent session $t>0$, the model only has access to the current training set $\mathcal{D}_{\mathrm{train}}^{t}$, which contains $N$ novel classes with $K$ labeled samples per class, i.e., an $N$-way $K$-shot setting.

At the testing phase of the $t$-th session, the model is evaluated over all previously encountered categories. The cumulative class set is defined as
$
\mathcal{Y}_t = \bigcup_{j=0}^{t} \mathcal{C}_j,
$
and the corresponding test label space is
$
\mathcal{C}_{t}^{\mathrm{test}} = \mathcal{Y}_t.
$ For convenience, when considering the $t$-th incremental session, we denote by
$
C_{old}=|\mathcal{Y}_{t-1}|
$
the number of previously seen classes before session $t$.
Compared with conventional FSCIL, G-FSCIL is characterized by a much smaller base-class ratio $|\mathcal{C}_0|/|\mathcal{Y}_T|$ and a larger number of incremental sessions, making the task substantially more challenging. 
A detailed comparison is given in Table~\ref{tab:gfscil_protocol}.

\begin{table}[t]
\centering
\caption{Comparison between conventional FSCIL and the proposed G-FSCIL protocol.}
\setlength{\tabcolsep}{2pt}
\scalebox{0.75}{
\begin{tabular}{lcccc}
\toprule
Dataset & Setting & Base Classes & Incremental Split & \#Total Sessions \\
\midrule
\multirow{2}{*}{CIFAR100 \cite{cifar-100}} 
& Conventional FSCIL & 60 & 5-way 5-shot & 9 \\
& G-FSCIL & 5 & 5-way 5-shot & 20 \\
\midrule
\multirow{2}{*}{CUB200 \cite{cub200}} 
& Conventional FSCIL & 100 & 10-way 5-shot & 11 \\
& G-FSCIL & 10 & 10-way 5-shot & 20 \\
\midrule
\multirow{2}{*}{ImageNet-R \cite{imagenet-r}} 
& Conventional FSCIL & 100 & 10-way 5-shot & 11 \\
& G-FSCIL & 10 & 10-way 5-shot & 20 \\
\bottomrule
\end{tabular}}
\label{tab:gfscil_protocol}
\end{table}

\noindent\textbf{ASP Backbone.}
We build our method upon ASP~\cite{liu2024few}, a representative prompt-based FSCIL framework. 
ASP freezes a pre-trained ViT backbone and inserts learnable prompts into the attention blocks, so as to preserve the generalization ability of the pre-trained model under few-shot incremental learning. 
At the $l$-th layer, the prompt is decomposed into a task-invariant part and a task-specific part, i.e., $p^l=[p_I^l; p_S^l]$. 
The task-specific prompt is constructed by combining a global prompt memory with an instance-aware prompt generated by a prompt encoder:
\begin{equation}
p_S^l=\alpha_p\,p_{avg}^l+(1-\alpha_p)\,f_\mu([p_I^l;f_\theta(x)]),
\end{equation}
where $f_\theta(\cdot)$ denotes the frozen backbone feature extractor, $f_\mu(\cdot)$ is the prompt encoder, $p_{avg}^l$ is the prompt memory updated across sessions, and $\alpha_p\in[0,1]$ is a balancing coefficient that controls the contribution of the prompt memory and the instance-aware prompt.

For classification, ASP uses a prototype-based classifier, where each class prototype is the mean prompt-enhanced feature, $c_k=\frac{1}{N_k}\sum_{y_i=k} f_{\theta,p}(x_i)$,
where $f_{\theta,p}(x_i)$ denotes the prompt-enhanced feature of sample $x_i$. 
Prototypes of newly encountered classes are appended incrementally during FSCIL. The prompt encoder is optimized with an information-bottleneck objective $\mathcal{L}_{IB}$ and anchor regularization $\mathcal{L}_{anc}$~\cite{liu2024few}:
\begin{equation}
\mathcal{L}^{real}_{ASP}
=
\mathcal{L}_{IB}
+
\lambda_{anc}\mathcal{L}_{anc}.
\end{equation}

\subsection{Synthetic Class Knowledge Construction}

{\looseness=-1
To alleviate the severe lack of labeled data in G-FSCIL, we employ a frozen pre-trained conditional latent diffusion model to construct class-specific synthetic knowledge without re-training the whole generator. 
Given the current session $t$, let $\mathcal{D}^{c}_{real}\subset \mathcal{D}^{t}_{train}$ denote the real image set of class $c\in\mathcal{C}_t$. 
For each class, we optimize a learnable inversion embedding $v_c$, which serves as a compact and reusable condition representation of the class distribution. 
Specifically, the embedding is obtained by class inversion:
\par}
\begin{equation}
v_c^{*}=\arg\min_{v_c}
\mathbb{E}_{\substack{z\sim \mathcal{E}(\mathcal{D}^{c}_{real})\\
\epsilon\sim\mathcal{N}(0,\mathbf{I}),\, t}}
\left[
\left\|\epsilon-
\epsilon_\Theta\!\bigl(
\substack{\sqrt{\alpha_t}z+\sqrt{1-\alpha_t}\epsilon,\\
t,\,[e_\theta(\tau);v_c]}
\bigr)\right\|_2^2
\right],
\end{equation}
where $\mathcal{E}(\cdot)$ is the latent encoder, $e_\theta(\tau)$ denotes the text embedding of a fixed prompt template $\tau$, $\epsilon_\Theta(\cdot)$ is the denoising network, and $\alpha_t$ is the diffusion coefficient at timestep $t$.

After class inversion, the learned embedding $v_c^{*}$ is used to guide the frozen latent diffusion model for conditional synthesis. 
For each class $c$, synthetic samples are generated by running a DDIM-based conditional sampler with unconditional guidance under the condition $[e_\theta(\tau);v_c^{*}]$, followed by decoding the generated latent through the pretrained decoder $\mathrm{Dec}(\cdot)$. 
Formally, the $j$-th synthetic sample of class $c$ is written as
\begin{equation}
\hat{x}_j^{(c)} = \mathrm{Dec}\!\left(\hat{\mathbf{z}}_0^{(c,j)}\right),
\qquad \hat{\mathbf{z}}_0^{(c,j)} \sim p_{\Theta}\bigl(\mathbf{z}_0 \mid [e_\theta(\tau);v_c^{*}]\bigr),
\end{equation}
where $p_{\Theta}$ denotes the conditional DDIM sampling process induced by the frozen latent diffusion model. 
In this way, we construct a candidate synthetic pool for session $t$:
\begin{equation}
\tilde{\mathcal{D}}^{t}_{syn}
=
\bigcup_{c\in\mathcal{C}_t}
\{(\hat{x}^{(c)}_j,c)\}_{j=1}^{M_{\mathrm{syn}}},
\end{equation}
where $M_{\mathrm{syn}}$ is the number of candidate synthetic samples generated for each class. This design ensures that synthetic data are not pre-generated for later sessions, but generated on demand from stored embeddings of previously seen classes.

\subsection{Learning to Curate Trustworthy Synthetic Knowledge}

Compared with conventional FSCIL, our G-FSCIL provides only limited labeled data in the base session, making the learned class prototypes more vulnerable to bias and instability.
Synthetic samples can provide additional within-class variation for observed classes, but low-quality generations may also distort the already formed feature space. 
We therefore treat synthetic data as a \emph{candidate knowledge pool} and learn to curate it via a knowledge curation strategy that assigns learnable soft weights to synthetic samples and optimizes them with semantic- and diversity-aware regularization.
The resulting curation behavior is further distilled into a transferable trust policy, enabling reliable synthetic knowledge to be reused in subsequent incremental sessions.

\noindent\textbf{Adaptive synthetic curation.}
Instead of assigning all generated samples equal importance, we associate each synthetic sample $\hat{x}_i\in\tilde{\mathcal{D}}^{0}_{syn}$ with a learnable scalar $w_i$, which is transformed into a soft selection weight $m_i=\sigma(\,w_i)$. Here $\sigma(\cdot)$ is the sigmoid function. 
The weighted synthetic classification loss is defined as
\begin{equation}
\mathcal{L}_{cls}^{syn}
=
\frac{1}{\sum_i m_i}
\sum_i m_i\, CE(\hat{x}_i,y_i).
\end{equation}
where $CE(\hat{x}_i,y_i)$ denotes the cross-entropy loss between the prediction of $\hat{x}_i$ and its label $y_i$.

To favor synthetic samples that are semantically aligned with their target classes while providing complementary intra-class variation, we measure both the semantic alignment and the diversity of each synthetic sample $\hat{x}_i$. Specifically, the semantic alignment score is defined as \(s_i=\cos\!\bigl(f_{\mathrm{img}}(\hat{x}_i),\, f_{\mathrm{text}}(y_i)\bigr)\),
where $f_{\mathrm{img}}(\cdot)$ and $f_{\mathrm{text}}(\cdot)$ denote the CLIP image and text encoders, respectively. The diversity score is defined as
\begin{equation}
d_i=
\frac{1}{|\mathcal{N}_i|}
\sum_{j\in\mathcal{N}_i}
\left\|
f_{\mathrm{img}}(\hat{x}_i)-f_{\mathrm{img}}(\hat{x}_j)
\right\|_2,
\end{equation}
where $\mathcal{N}_i$ contains the $K_N$ nearest same-class candidates to $\hat{x}_i$ in CLIP image space. Given the soft selection weights $m_i$, we define the curation regularizer as
\begin{equation}
\mathcal{L}_{cur}
=
-\frac{1}{M_{\mathrm{syn}}}\sum_{i=1}^{M_{\mathrm{syn}}} m_i(\hat{s}_i+\lambda_{div}\hat{d}_i)
+
\beta
\left|
\frac{1}{M_{\mathrm{syn}}}\sum_{i=1}^{M_{\mathrm{syn}}} m_i-\rho_{\mathrm{sel}}
\right|,
\end{equation}
where $\hat{s}_i$ and $\hat{d}_i$ are the normalized semantic alignment and diversity scores, respectively, $\rho_{\mathrm{sel}}$ denotes the target selection fraction within each fixed class-specific synthetic pool, rather than the synthetic-to-real training ratio, and $\lambda_{\mathrm{div}}$ and $\beta$ control the diversity and selection-ratio regularization terms.

\noindent\textbf{Transferable Selection Policy.}
The base-session adaptive mask is effective for synthetic curation but cannot be directly reused in later sessions, while re-learning it under limited supervision is inefficient and unreliable. We therefore distill its curation behavior into a lightweight policy scorer, using the mask as a teacher and compact sample-level cues as student inputs.

For each synthetic sample $\hat{x}_i$, we construct a quality descriptor
$\mathbf{r}_i=[s_i,p_i,d_i]^\top$, where $s_i$ and $d_i$ denote the semantic alignment and diversity scores defined above, respectively. The prototype-consistency score $p_i$ is computed as $p_i=\cos\!\left(f_{\theta,p}(\hat{x}_i),\mu_{y_i}\right)$,
where $\mu_{y_i}$ is the prototype of class $y_i$ computed from its real samples $\mathcal{D}_{\mathrm{real}}^{y_i}$.
The policy scorer \(g_\phi(\cdot)\) further maps the descriptor \(\mathbf{r}_i\) to a soft selection score \(m_i^{trust}=\sigma(g_\phi(\mathbf{r}_i))\),
where $g_\phi(\cdot)$ denotes a lightweight MLP parameterized by $\phi$, and $\sigma(\cdot)$ is the sigmoid function that maps the output to $(0,1)$.

During the base session, the policy scorer is trained to mimic the soft selection behavior of the teacher mask. Let $m_i^{tea}$ denote the teacher weight assigned to $\hat{x}_i$. We optimize the student by
\begin{equation}
\mathcal{L}_{policy}
=
\frac{1}{M_{\mathrm{syn}}}
\sum_{i=1}^{M_{\mathrm{syn}}}
\left\|
m_i^{trust}
-
\operatorname{sg}(m_i^{tea})
\right\|_2^2,
\end{equation}
where $\operatorname{sg}(\cdot)$ denotes the stop-gradient operator. In this way, the student inherits the synthetic curation preference induced by the teacher, without affecting the optimization of the teacher mask itself.

The overall base-session objective can be defined as
\begin{equation}
\mathcal{L}_{Base}
=
\mathcal{L}^{real}_{ASP}
+
\mathcal{L}_{cls}^{syn}
+
\mathcal{L}_{cur}
+
\mathcal{L}_{policy}.
\end{equation}

After the base session, we discard the teacher mask and retain only the policy scorer $g_{\phi}(\cdot)$ to evaluate synthetic candidates in later sessions. Thus, synthetic trustworthiness is inherited from the base-session curation policy rather than re-learned from few-shot data, providing a reusable selection prior for incremental adaptation.

\subsection{Boundary-Stable Incremental Adaptation}

In G-FSCIL, the challenge of incremental adaptation lies not only in limited supervision, but also in how to stably integrate few-shot novel classes into a feature space already occupied by previously learned classes. Under such limited supervision, novel-class prototypes are often estimated unreliably, which in turn induces boundary conflicts with old classes during subsequent adaptation. To address this issue, we propose a \emph{boundary-stable incremental adaptation} strategy. It first performs \emph{synthetic-informed prototype initialization} from trustworthy synthetic knowledge, and then applies \emph{bidirectional boundary calibration} to suppress high-risk interference between old and novel classes. In this way, the model can stably adapt to novel classes using sparse real samples together with reliable synthetic knowledge selected by the learned policy.

\noindent\textbf{Synthetic-informed prototype initialization.} At the beginning of each incremental session $t$, we first synthesize class-wise candidate samples for all seen classes using the saved class inversion embeddings $
\mathcal{C}_{\le t}=\mathcal{Y}_t=\bigcup_{j=0}^{t}\mathcal{C}_j.$

For each class \(c\in\mathcal{C}_{\le t}\), we retrieve its cached synthetic candidate pool \(\tilde{\mathcal{D}}_{syn}^{t,c}=\{(\hat{x}^{(c)}_j,c)\}_{j=1}^{M_{\mathrm{syn}}}\).
Rather than using these cached candidates directly, we evaluate them with the transferable selection policy learned in the base session and retain \(\bar{\mathcal{D}}_{syn}^{t,c}=\operatorname{TopK}\!\bigl(\tilde{\mathcal{D}}_{syn}^{t,c}\bigr)\), with \(K=50\), for each class.
Based on the filtered class-wise pools, we further define the current-class and old-class synthetic sets as
\begin{equation}
\bar{\mathcal{D}}_{cur}^{t}
=
\bigcup_{c\in\mathcal{C}_t}\bar{\mathcal{D}}_{syn}^{t,c},
\qquad
\bar{\mathcal{D}}_{old}^{t}
=
\bigcup_{c\in\mathcal{C}_{\le t-1}}\bar{\mathcal{D}}_{syn}^{t,c}.
\end{equation}
All synthetic supports and probes used in the current session are sampled only from these filtered high-quality pools.

Before gradient-based adaptation begins, we use these trustworthy synthetic candidates to stabilize the initialization of current novel classes. For each novel class $c\in\mathcal{C}_t$, let
$
D_{real}^{c}\subset\mathcal{D}_{train}^{t}
$
denote its few-shot real sample set. We first compute the real prototype by averaging the current-session real features:
\begin{equation}
\mu_{c}^{real}
=
\frac{1}{|D_{real}^{c}|}
\sum_{x_i\in D_{real}^{c}} f_{\theta,p}(x_i).
\end{equation}
Following ASP, the prompt memory is updated using the inherited EMA rule on current-session real samples only.

\begin{table*}[t]
\centering
\caption{Comparison with SOTA methods on ImageNet-R under the G-FSCIL protocol.}
\label{tab:imagenetr}

\footnotesize
\setlength{\tabcolsep}{2pt}
\renewcommand{\arraystretch}{1.0}
\scalebox{1.0}{
\begin{tabular}{lcccccccccccccccccccc@{\hspace{4pt}}c@{\hspace{6pt}}cc}
\toprule
\multirow{2}{*}{Method} & \multicolumn{20}{c}{Acc. in each session $\uparrow$} & \multirow{2}{*}{PD $\downarrow$} & \multirow{2}{*}{HAcc $\uparrow$} & \multirow{2}{*}{AvgAcc $\uparrow$} \\
\cmidrule(lr){2-21}
& 0 & 1 & 2 & 3 & 4 & 5 & 6 & 7 & 8 & 9 & 10 & 11 & 12 & 13 & 14 & 15 & 16 & 17 & 18 & 19 & & & \\
\midrule

Coda-Prompt~\cite{smith2023codaprompt}
& \textbf{93.8} & 26.0 & 16.3 & 11.8 & 5.0 & 7.1 & 3.0 & 3.0 & 3.7 & 5.3
& 3.5 & 1.6 & 2.3 & 1.9 & 1.7 & 2.3 & 1.6 & 1.6 & 1.2 & 1.6
& 92.2 & 0.0 & 14.8 \\

DualPrompt~\cite{wang2022dualprompt}
& 27.4 & 13.5 & 10.1 & 4.6 & 2.7 & 1.9 & 1.4 & 1.3 & 1.3 & 0.7
& 0.5 & 0.6 & 0.4 & 0.4 & 0.3 & 0.3 & 0.4 & 0.4 & 0.3 & 0.3
& 27.1 & 4.0 & 3.3 \\

L2P~\cite{wang2022l2p}
& 86.4 & 63.4 & 37.7 & 21.7 & 16.7 & 11.9 & 9.8 & 6.6 & 6.7 & 4.2
& 3.3 & 3.5 & 2.5 & 3.1 & 2.5 & 3.5 & 4.4 & 3.3 & 2.3 & 2.1
& 84.3 & 4.4 & 14.6 \\
DDGR~\cite{gao2023ddgr}
& 38.1 & 20.6 & 11.0 & 9.2 & 7.3 & 6.1 & 5.8 & 4.5 & 3.6 & 3.0
& 2.6 & 2.1 & 1.9 & 1.8 & 1.8 & 1.6 & 1.8 & 1.6 & 1.9 & 1.8
& 36.3 & 2.4 & 5.7 \\
\midrule

LGSP-Prompt~\cite{jiang2025lgsp}
& 61.0 & 50.8 & 34.7 & 21.9 & 16.6 & 14.1 & 12.2 & 9.9 & 6.7 & 6.8
& 4.3 & 4.4 & 4.0 & 2.7 & 3.0 & 2.7 & 2.5 & 1.9 & 2.0 & 1.9
& 59.1 & 0.0 & 13.5 \\

TEEN~\cite{wang2023few}
& 35.7 & 13.7 & 9.1 & 7.4 & 6.3 & 5.4 & 5.0 & 4.8 & 3.7 & 3.5
& 3.0 & 2.9 & 2.6 & 2.3 & 2.2 & 2.0 & 1.9 & 1.7 & 1.9 & 1.7
& 34.0 & 0.6 & 4.5 \\

SEC-Prompt~\cite{liu2025sec}
& 88.5 & 67.4 & 59.8 & 54.6 & 52.8 & 51.8 & 50.8 & 47.3 & 46.2 & 45.0
& 44.3 & 44.7 & 41.9 & 41.5 & 40.6 & 40.5 & 40.5 & 40.9 & 40.5 & 39.7
& 48.8 & 39.1 & 48.9 \\

ASP~\cite{liu2024few}
& 89.0 & 80.2 & 69.7 & 68.9 & 69.0 & 69.2 & 69.3 & 65.6 & 63.1 & 61.3
& 59.6 & 57.8 & 57.5 & 57.0 & 55.8 & 55.8 & 55.4 & 54.9 & 54.4 & 53.2
& 35.8 & 63.2 & 63.7 \\

\midrule

\textbf{Ours}
& 92.6 & \textbf{81.8} & \textbf{79.1} & \textbf{75.5} & \textbf{76.4} & \textbf{74.1} & \textbf{75.0} & \textbf{72.4} & \textbf{70.9} & \textbf{68.0}
& \textbf{67.3} & \textbf{65.8} & \textbf{66.9} & \textbf{65.3} & \textbf{65.1} & \textbf{63.2} & \textbf{61.6} & \textbf{62.2} & \textbf{65.9} & \textbf{64.6}
& \textbf{28.0} & \textbf{65.4} & \textbf{70.7} \\

\bottomrule
\end{tabular}}
\end{table*}

To alleviate the bias caused by sparse supervision, we further construct a trustworthy synthetic support set
$
\mathcal{S}_{c}^{proto}\subset\bar{\mathcal{D}}_{cur}^{t}
$
for each novel class using the learned selection policy. The corresponding synthetic prototype is computed by trust-weighted averaging:
\begin{equation}
\mu_c^{syn}
=
\frac{
\sum_{\hat{x}_j^{syn}\in\mathcal{S}_c^{proto}}
m_j^{trust}\, f_{\theta,p}(\hat{x}_j^{syn})
}{
\sum_{\hat{x}_j^{syn}\in\mathcal{S}_c^{proto}}
m_j^{trust}
},
\end{equation}
where $m_j^{trust}$ is the trust score predicted by the transferable policy scorer. We then fuse the real and synthetic prototypes to obtain a safer initialization for class $c$:
\begin{equation}
w_c^{new}
=
\frac{\mu_c^{real}+\rho_c\,\mu_c^{syn}}{1+\rho_c},
\end{equation}
where $\rho_c$ controls the contribution of trustworthy synthetic supports. Old-class weights are inherited from the stored prototype memory, while $w_c^{new}$ is used as the classifier weight for each novel class. By stabilizing the entry point of each novel class into the classifier space, this step reduces prototype bias and alleviates unnecessary feature drift during subsequent adaptation. It is worth noting that during incremental adaptation, only the prompt encoder is fine-tuned on current novel-class real samples, while the backbone remains frozen.

\noindent\textbf{Bidirectional boundary calibration.}
Although synthetic-informed prototype initialization provides a more reliable starting point for novel classes, few-shot adaptation can still introduce local conflicts between old and novel classes. Current novel samples may be biased toward semantically similar old classes, while adaptation to new classes can disturb the decision regions of previously learned classes. We therefore perform a local boundary refinement that selectively enlarges the margins for risky interactions in the novel-to-old and old-to-novel directions.

During session $t$, the cross-entropy loss is computed only on real samples from the current session, while synthetic samples are used only for prototype initialization and reverse calibration. To focus the classification objective on within-session discrimination, old-class logits are masked in the cross-entropy loss:

\begin{equation}
z^{ce}_{i,k}
=
\begin{cases}
-\infty, & k<C_{old},\\
z_{i,k}, & k\ge C_{old},
\end{cases}
\qquad
\mathcal{L}_{ce}
=
\frac{1}{|\mathcal{D}_{train}^{t}|}\sum_i CE(z_i^{ce},y_i),
\end{equation}
where $C_{old}$
denotes the number of previously seen classes before the current session. $|\mathcal{D}_{train}^{t}|$ denotes the number of real training samples in the current session. $CE(z_i^{ce},y_i)$ denotes the cross-entropy loss between the prediction of $z_i^{ce}$ and its label $y_i$.

\emph{Forward calibration from current novel samples to old classes.}
For a current real sample \(x_i\) with label \(y_i\in\mathcal{C}_t\), we denote its target novel-class logit and strongest old-class competitor by \(z_i^{n}=z_{i,y_i}\) and \(z_i^{o}=\max_{k<C_{old}}z_{i,k}\), respectively.
A boundary violation occurs when the competing old-class response $z_i^{o}$ approaches or exceeds $z_i^{n}$. We therefore enlarge the margin between them using a margin $\delta$, yielding the forward calibration loss
\begin{equation}
\mathcal{L}_{fwd}
=
\frac{1}{|\mathcal{D}_{train}^{t}|}
\sum_{x_i\in\mathcal{D}_{train}^{t}}
\max\!\bigl(0,\; z_i^{o}-z_i^{n}+\delta\bigr),
\end{equation}

\emph{Reverse calibration from old probes to current novel classes.}
While the forward term prevents current novel samples from being absorbed by old classes, it does not prevent the current adaptation process from collectively invading old-class regions. To preserve previously learned decision regions, we introduce a reverse calibration term using trustworthy old-class probes sampled from $\bar{\mathcal{D}}_{old}^{t}$.

For an old probe \(\tilde{x}_j^{old}\) with source label \(\tilde{y}_j^{old}\), we define \(z_j^{old}=z_{j,\tilde{y}_j^{old}}\) and \(z_j^{new}=\max_{k\in\mathcal{C}_t}z_{j,k}\).

Here, $z_j^{old}$ is the classifier response to its source old class, while $z_j^{new}$ is the strongest response among current novel classes. We then enforce the reverse calibration loss
\begin{equation}
\mathcal{L}_{rev}
=
\frac{1}{M_{old}}
\sum_j
\max\!\bigl(0,\; z_j^{new}-z_j^{old}+\delta\bigr),
\end{equation}
where $M_{old}$ is the number of sampled old probes. This term explicitly discourages the adaptation process from shifting old-class samples into the decision regions of current novel classes.

Finally, the incremental objective is
\begin{equation}
\mathcal{L}_{inc}
=
\mathcal{L}_{ce}
+
\lambda_{fwd}\mathcal{L}_{fwd}
+
\lambda_{rev}\mathcal{L}_{rev},
\end{equation}
where $\lambda_{rev}$ and $\lambda_{fwd}$ are balancing parameters.

\section{Experiments}

\begin{table*}[t]
\centering
\caption{Comparison with SOTA methods on CIFAR100 under the G-FSCIL protocol.}
\label{tab:main_results}

\footnotesize
\setlength{\tabcolsep}{2pt}
\renewcommand{\arraystretch}{1.0}

\scalebox{1.0}{
\begin{tabular}{lcccccccccccccccccccc@{\hspace{4pt}}c@{\hspace{6pt}}cc}
\toprule
\multirow{2}{*}{Method} & \multicolumn{20}{c}{Acc. in each session $\uparrow$} & \multirow{2}{*}{PD $\downarrow$} & \multirow{2}{*}{HAcc $\uparrow$} & \multirow{2}{*}{AvgAcc $\uparrow$} \\
\cmidrule(lr){2-21}
& 0 & 1 & 2 & 3 & 4 & 5 & 6 & 7 & 8 & 9 & 10 & 11 & 12 & 13 & 14 & 15 & 16 & 17 & 18 & 19 & & & \\
\midrule

Coda-Prompt~\cite{smith2023codaprompt}
& 99.0 & 68.2 & 23.7 & 12.8 & 14.9 & 8.1 & 6.5 & 4.5 & 4.6 & 2.6
& 6.1 & 3.4 & 4.7 & 3.4 & 3.9 & 2.4 & 2.7 & 1.8 & 2.2 & 1.3
& 97.7 & 0.0 & 13.8 \\

DualPrompt~\cite{wang2022dualprompt}
& 99.2 & 83.3 & 69.4 & 54.3 & 45.0 & 37.2 & 32.5 & 26.6 & 22.6 & 20.5
& 17.9 & 14.2 & 12.6 & 10.9 & 9.3 & 8.0 & 7.1 & 6.1 & 5.2 & 4.5
& 94.7 & 0.4 & 29.3 \\

L2P~\cite{wang2022l2p}
& 96.8 & 77.1 & 57.5 & 45.5 & 40.2 & 26.1 & 17.1 & 11.0 & 7.6 & 8.2
& 7.8 & 5.7 & 5.2 & 6.1 & 3.2 & 5.3 & 4.9 & 3.0 & 3.7 & 3.2
& 93.6 & 6.5 & 21.8 \\

DDGR~\cite{gao2023ddgr}
& 84.0 & 44.2 & 26.5 & 14.7 & 15.7 & 9.4 & 7.1 & 8.8 & 7.2 & 6.4
& 6.7 & 5.2 & 6.2 & 4.5 & 4.1 & 3.7 & 3.6 & 3.4 & 2.7 & 3.5
& 80.5 & 1.0 & 13.4 \\
\midrule

LGSP-Prompt~\cite{jiang2025lgsp}
& 98.8 & 95.0 & 81.4 & 58.0 & 55.3 & 51.3 & 43.4 & 39.4 & 28.1 & 22.4
& 15.1 & 9.0 & 6.2 & 2.6 & 1.3 & 0.8 & 0.5 & 0.5 & 0.3 & 0.3
& 89.8 & 0.8 & 30.5 \\

TEEN~\cite{wang2023few}
& 86.6 & 49.6 & 34.5 & 25.9 & 22.6 & 19.8 & 18.2 & 16.4 & 14.4 & 13.4
& 13.2 & 12.4 & 12.0 & 11.2 & 10.9 & 10.3 & 9.9 & 9.6 & 9.0 & 8.7
& 77.9 & 11.9 & 20.4 \\

SEC-Prompt~\cite{liu2025sec}
& 99.2 & 96.1 & 91.0 & 88.2 & 87.1 & 83.3 & 81.9 & 79.0 & 78.3 & 76.5
& 75.1 & 72.9 & 71.2 & 71.1 & 69.5 & 68.7 & 68.5 & 68.8 & 68.6 & 68.0
& 31.2 & 68.8 & 78.2 \\

ASP~\cite{liu2024few}
& 99.0 & 94.6 & 88.3 & 85.0 & 84.0 & 79.2 & 76.4 & 74.7 & 73.6 & 71.4
& 71.1 & 69.3 & 66.7 & 66.3 & 64.2 & 64.6 & 64.3 & 64.5 & 63.3 & 62.4
& 36.6 & 72.4 & 74.1 \\

\midrule

\textbf{Ours}
& \textbf{99.4} & \textbf{98.6} & \textbf{95.1} & \textbf{93.3} & \textbf{93.1} & \textbf{90.1} & \textbf{88.7} & \textbf{87.7} & \textbf{87.4} & \textbf{86.1}
& \textbf{85.2} & \textbf{83.8} & \textbf{82.7} & \textbf{82.7} & \textbf{81.3} & \textbf{81.2} & \textbf{80.9} & \textbf{81.2} & \textbf{80.2} & \textbf{79.0}
& \textbf{20.4} & \textbf{84.2} & \textbf{86.9} \\

\bottomrule
\end{tabular}}
\end{table*}

\subsection{Experimental Setup}

\stitle{Datasets.}
We evaluate our method on CIFAR100 \cite{cifar-100}, CUB200-2011 \cite{cub200}, and ImageNet-R \cite{imagenet-r} under the proposed G-FSCIL protocol. In contrast to conventional FSCIL settings with a large base session, our protocol adopts a significantly smaller base-class pool, resulting in substantially weaker initial priors and a more realistic yet challenging evaluation scenario. As detailed in Table \ref{tab:gfscil_protocol}, CIFAR100 is initialized with 5 base classes under a 5-way 5-shot incremental setting, while CUB200 and ImageNet-R use 10 base classes with a 10-way 5-shot configuration. 


\stitle{Evaluation protocol.}
After the \(t\)-th session, we report the Top-1 accuracy \(A_t\) over all classes observed so far, i.e., \(\mathcal{C}_0 \cup \cdots \cup \mathcal{C}_t\). We further report the average incremental accuracy
$
A_{\text{avg}}=\frac{1}{T+1}\sum_{t=0}^{T} A_t,
$
which measures the overall performance across the whole learning process. To quantify forgetting, we use the performance drop
$\mathrm{PD}=A_0-A_T,$
where \(A_0\) and \(A_T\) denote the accuracies after the base session and the last session, respectively. Following prior FSCIL practice \cite{liu2024few}, we also report the harmonic accuracy at the last session,
$A_h=\frac{2A_oA_n}{A_o+A_n},$
where \(A_o\) is the final accuracy on base classes and \(A_n\) is the final accuracy on incremental classes.

\stitle{Implementation details.}
We adopt a ViT-B/16 backbone \cite{vit-b} pretrained on ImageNet-1K \cite{imagenet} for all experiments. All input images are resized to \(224 \times 224\) to align with the input resolution of the ViT backbone and the synthetic generation pipeline. Synthetic samples are produced with a latent diffusion model pretrained on LAION-400M~\cite{schuhmann2021laion400m,rombach2022high}. For each class, we optimize a class-specific condition embedding via class inversion for 5 rounds with 50 steps, and generate \(M_{\mathrm{syn}} = 200\) candidate samples. The base session is trained with both real and synthetic samples for 10 epochs, with learning rates of 0.01, 0.05, and 0.02 on CIFAR100, CUB200, and ImageNet-R, respectively. The target within-pool selection fraction is dataset-specific, with \(\rho_{\mathrm{sel}}=0.01\) for CIFAR100 and ImageNet-R, and \(\rho_{\mathrm{sel}}=0.001\) for CUB200. Synthetic trustworthiness is estimated by an MLP-based policy scorer. During incremental sessions, optimization is performed using current real samples and old synthetic samples, while novel synthetic samples are used exclusively for prototype enhancement. The classifier head is frozen during incremental fine-tuning. Each incremental session is trained for 10 epochs with a learning rate of 0.01. We set the prototype fusion coefficient to \(\rho_c = 0.1\) and the calibration margin to \(\delta = 0.1\).

\stitle{Baseline methods.}
We compare our method with representative approaches from three categories: (1) general continual learning methods, including prompt-based approaches L2P~\cite{wang2022l2p}, DualPrompt~\cite{wang2022dualprompt}, and CODA-Prompt~\cite{smith2023codaprompt}, as well as the diffusion-based generative replay method DDGR~\cite{gao2023ddgr}; (2) FSCIL-oriented prompt-based methods, including LGSP-Prompt~\cite{jiang2025lgsp}, SEC-Prompt~\cite{liu2025sec}, and ASP~\cite{liu2024few}; and (3) TEEN~\cite{wang2023few}, a representative FSCIL method without prompt learning.

\subsection{Benchmark Comparisons}

Under G-FSCIL, Tables~\ref{tab:imagenetr} and~\ref{tab:main_results} compare representative FSCIL baselines by session-wise accuracy and three evaluation metrics on ImageNet-R and CIFAR100, respectively, while Fig.~\ref{fig: SOTA_CUB} presents CUB200 Top-1 accuracy \(A_t\) curves.

\begingroup
\setlength{\intextsep}{6pt plus 1pt minus 1pt}
\setlength{\belowcaptionskip}{4pt}
\begin{figure}[t]
   \centering
   \includegraphics[width=\linewidth]{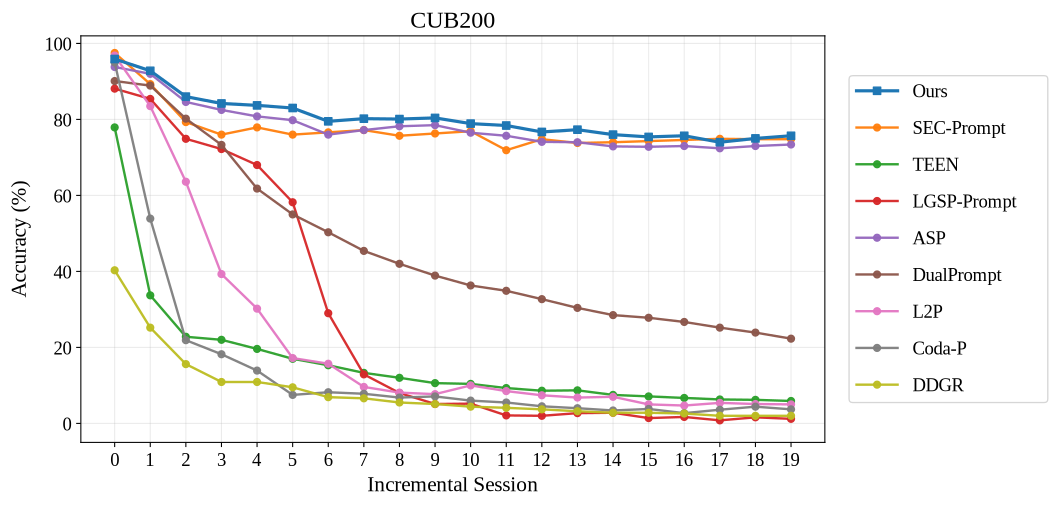}
   
    \caption{Comparison with SOTA methods on CUB200.}
   \Description{Line chart of Top-1 accuracy over 20 CUB200 sessions for nine methods. The proposed method remains highest or near highest throughout the sequence and ends around 76 percent, while several baselines decline sharply.}
   \label{fig: SOTA_CUB}
  
\end{figure}
\endgroup
Our method achieves the best performance on both ImageNet-R and CIFAR100. On ImageNet-R, it improves AvgAcc to 70.7 and HAcc to 65.4, while reducing PD to 28.0. These results demonstrate a clear advantage in balancing stability and plasticity under the low-base setting.
On CIFAR100, our method further achieves an AvgAcc of 86.9, an HAcc of 84.2, and a PD of 20.4, surpassing the strongest baseline SEC-Prompt by margins of 8.7, 15.4, and 10.8, respectively.
As shown in Fig.~\ref{fig: SOTA_CUB}, our method consistently maintains higher accuracy across sessions with a flatter degradation trend, indicating stronger resistance to long-term forgetting. This advantage becomes more pronounced in the middle and later stages, where accumulated old-new interference is more severe.
Overall, these results confirm the effectiveness of leveraging trustworthy synthetic knowledge under weak base priors, yielding more reliable prototype initialization and more stable boundary preservation throughout incremental learning.

\subsection{Ablation and Sensitivity Analysis}

\stitle{Component Ablation.}
Table~\ref{tab:component_ablation} evaluates the contributions of trustworthy synthetic knowledge curation (TSKC) and boundary-stable incremental adaptation (BSIA). TSKC alone consistently improves the baseline on both CIFAR100 and CUB200, showing that semantically aligned and diverse synthetic samples are already beneficial under weak base priors. In contrast, BSIA alone brings limited gains on CIFAR100 and is less stable on CUB200, indicating that its prototype and boundary calibration depends on a reliable base representation. Combining both components yields the best performance, as TSKC improves the base feature geometry and enables BSIA to integrate novel classes more reliably while reducing old--new interference.

\setlength{\tabcolsep}{2.2mm}
\renewcommand\arraystretch{1.1}

\begin{table}[t]
\centering
\small
\setlength{\tabcolsep}{3.5pt}

\renewcommand{\arraystretch}{0.90}
\caption{Grouped ablation study on CIFAR100 and CUB200.}
\label{tab:component_ablation}
\begin{tabular}{cc|ccc|ccc}
\toprule
\multicolumn{2}{c|}{\textbf{Modules}} 
& \multicolumn{3}{c|}{\textbf{CIFAR100}} 
& \multicolumn{3}{c}{\textbf{CUB200}} \\
\cmidrule(lr){1-2} \cmidrule(lr){3-5} \cmidrule(lr){6-8}
TSKC & BSIA
& AvgAcc $\uparrow$ & HAcc $\uparrow$ & PD $\downarrow$
& AvgAcc $\uparrow$ & HAcc $\uparrow$ & PD $\downarrow$ \\
\midrule

 & 
& 74.14 & 72.40 & 36.64
& 78.06 & 79.82 & 20.50 \\

 & \checkmark 
& 76.49 & 74.41 & 31.44
& 64.61 & 9.16 & 27.65 \\

\checkmark &  
& 78.99 & 77.92 & 30.65
& 78.36 & \textbf{81.80} & \textbf{19.12} \\

\checkmark & \checkmark 
& \textbf{86.89} & \textbf{84.23} & \textbf{20.37}
& \textbf{80.44} & 79.94 & 20.15 \\

\bottomrule
\end{tabular}
\end{table}

\begin{figure}[t]
   \centering
   \includegraphics[width=0.85\linewidth]{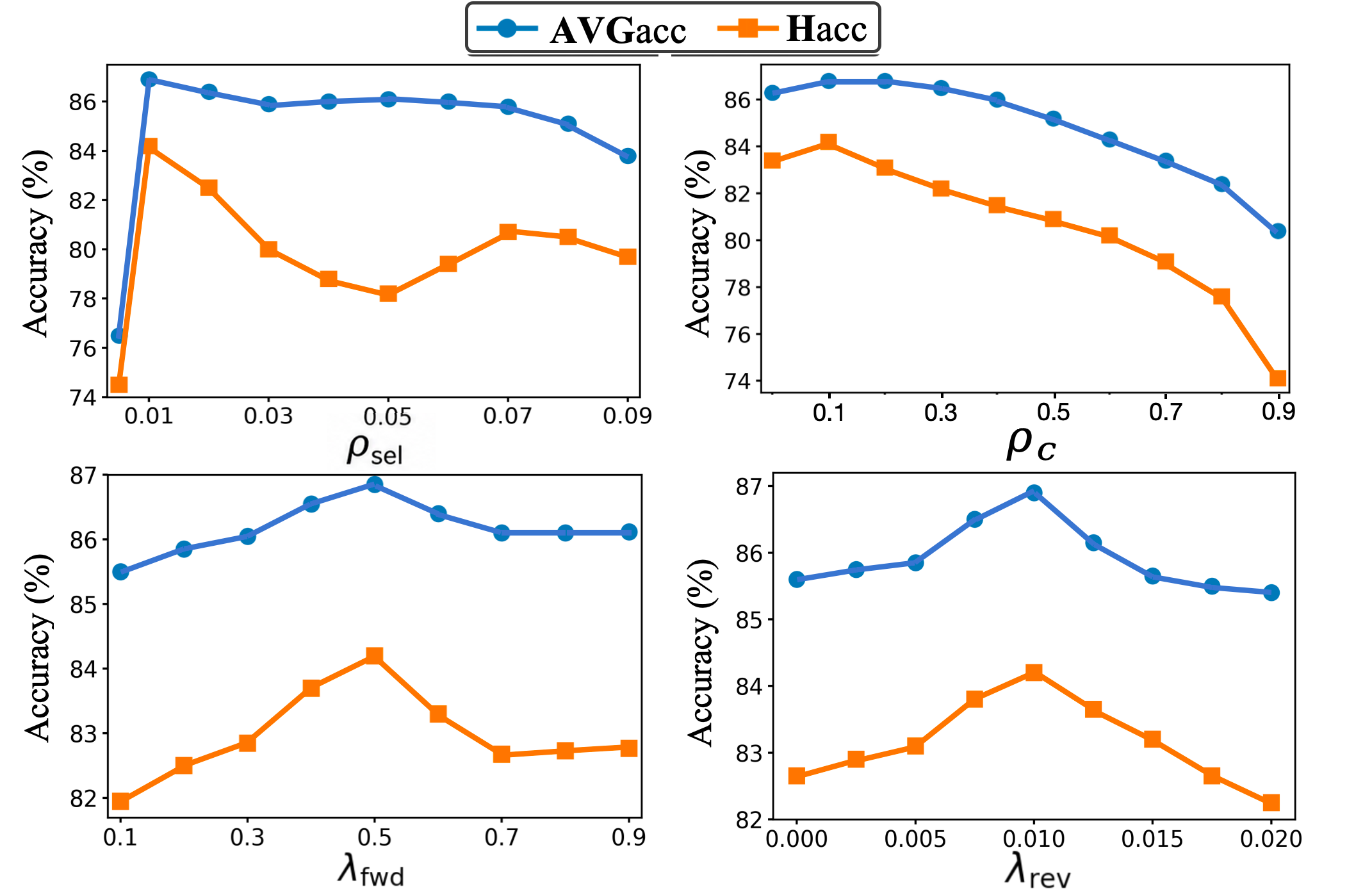}
    \caption{Sensitivity analysis of $\rho_{\mathrm{sel}}$, $\rho_c$, $\lambda_{fwd}$, and $\lambda_{rev}$.}
    \Description{Four line charts show average and harmonic accuracy as the within-pool selection fraction, prototype fusion coefficient, forward calibration weight, and reverse calibration weight vary. Performance peaks at a small selection fraction, declines as the fusion coefficient increases beyond a low value, and peaks at moderate forward and reverse calibration weights.}
   \label{fig: Hyper}
\end{figure}

\stitle{Hyperparameter Sensitivity.}
Fig.~\ref{fig: Hyper} evaluates the sensitivity to the selection sparsity $\rho_{\mathrm{sel}}$, prototype fusion weight $\rho_c$, and calibration weights $\lambda_{fwd}$ and $\lambda_{rev}$. A small $\rho_{\mathrm{sel}}$ yields the best performance, suggesting that conservative selection is beneficial for filtering unreliable synthetic candidates. Performance gradually decreases as $\rho_c$ increases, indicating that synthetic prototypes are more effective when used to complement rather than dominate real supervision. For both $\lambda_{fwd}$ and $\lambda_{rev}$, moderate values achieve the best results because small values provide insufficient boundary calibration, whereas large values impose overly strong constraints.

\subsection{Further Analysis}

\begin{figure}[t]
   \centering
   \includegraphics[width=0.9\linewidth]{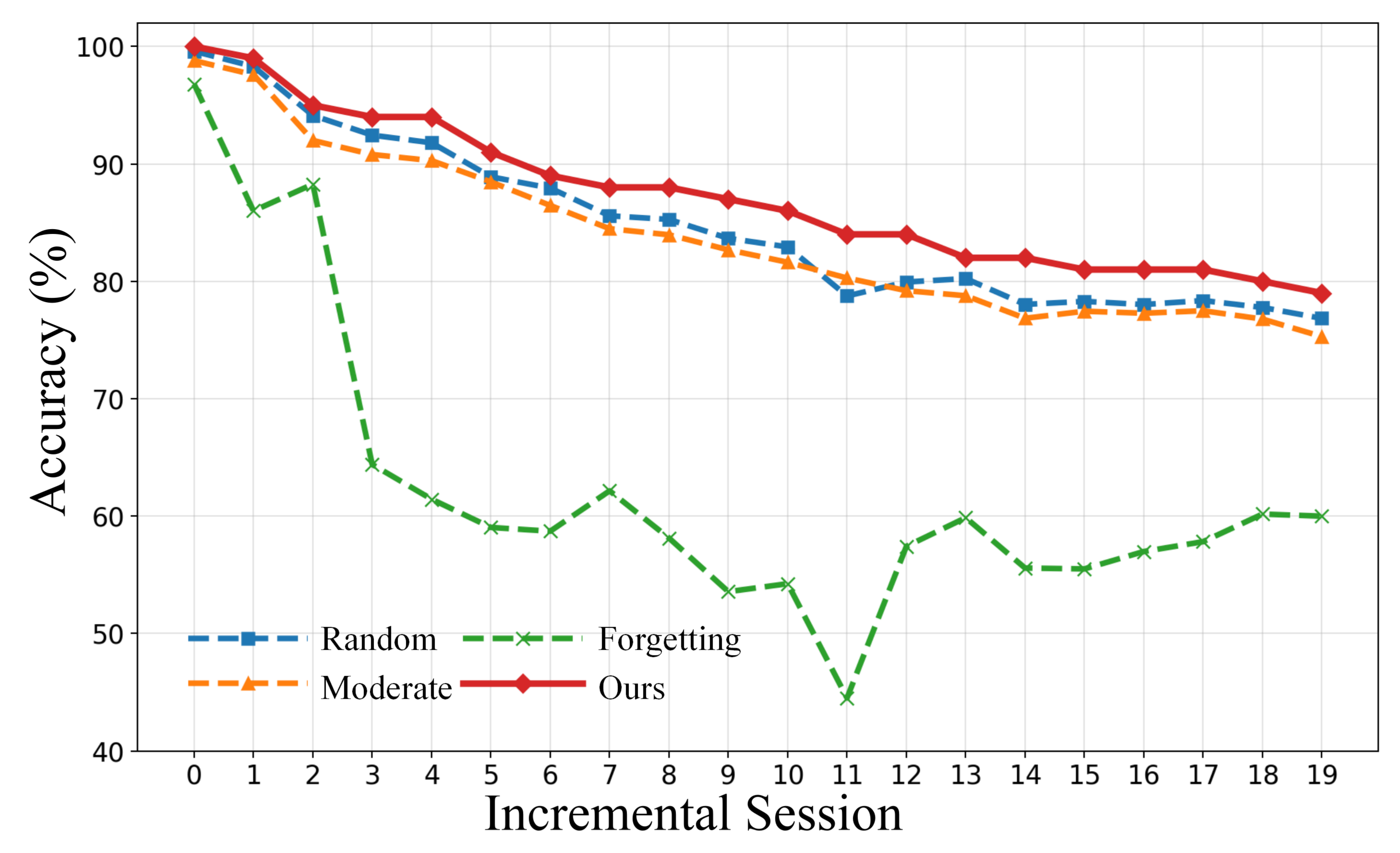}
   \caption{Comparison of different selection methods.}
   \Description{Line chart comparing classification accuracy over 20 incremental sessions for random, Moderate Coreset, forgetting-based, and learned selection. The learned policy is consistently best, while the forgetting-based strategy degrades substantially.}
   \label{fig:Selection Method}
\end{figure}

\stitle{Trustworthiness of the Learned Selection Policy.}
We compare our learned selection policy against three alternatives: random selection, Moderate Coreset~\cite{xia2023moderate}, and a forgetting-based strategy~\cite{toneva2018empirical}. As shown in Fig.~\ref{fig:Selection Method}, our policy consistently outperforms all three baselines throughout the incremental process, with especially clear gains in the middle and late sessions. This trend suggests that the learned selector captures transferable cues of synthetic trustworthiness, rather than relying on a session-specific heuristic. By contrast, the forgetting-based strategy performs substantially worse, suggesting that selection without explicit semantic reliability can damage the learned base geometry.

\begin{figure}[t]
   \centering
   \includegraphics[width=\linewidth]{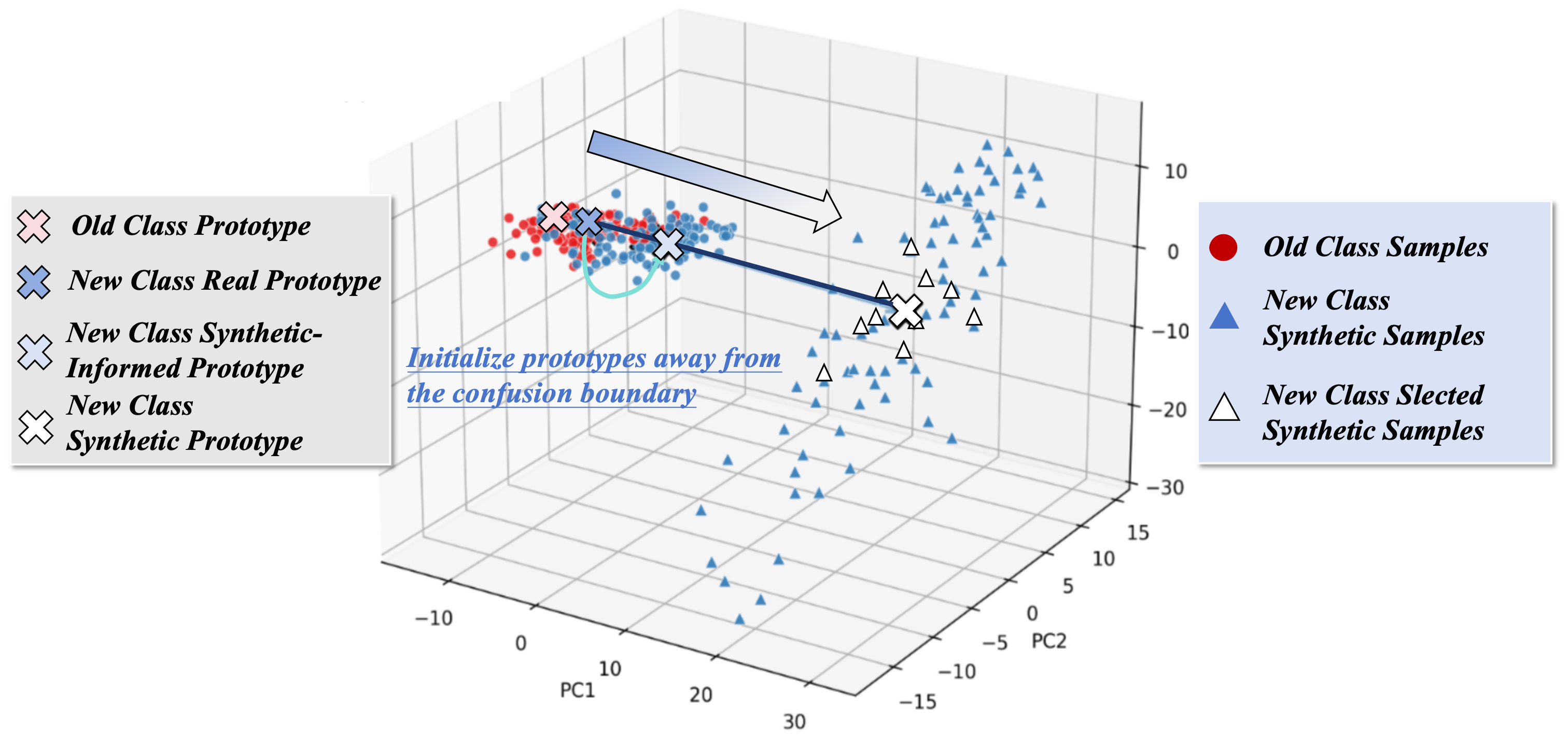}
   \caption{Visualization of synthetic-informed prototype initialization.}
   \Description{Three-dimensional PCA plot of old-class samples and novel-class synthetic samples. Selected synthetic samples move the novel-class prototype away from the old-class confusion boundary relative to the prototype computed only from few-shot real samples.}
   \label{fig:pca}
\end{figure}

\stitle{Geometric Effect of Prototype Initialization and Boundary Calibration.}
Fig.~\ref{fig:pca} shows that synthetic-informed prototype initialization moves the synthetic-informed novel prototype away from old-class high-density regions relative to the few-shot real prototype, indicating that trustworthy synthetic supports primarily correct the initial point of novel classes rather than merely enlarge the training set. This geometric correction reduces the chance that a new class is inserted directly into a confusion region.

\begin{figure}[t]
   \centering
   \includegraphics[width=\linewidth]{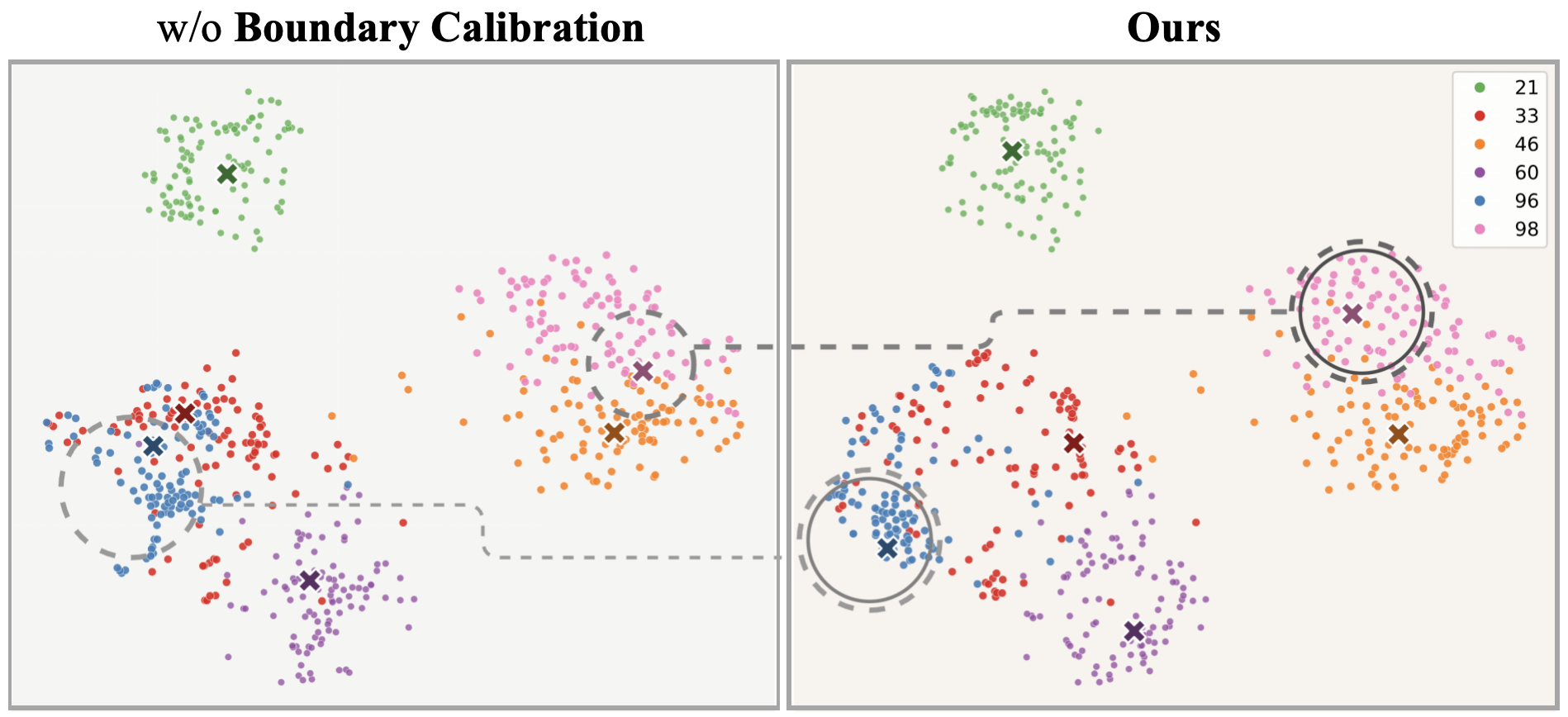}
   \caption{t-SNE visualization of Boundary Calibration.}
   \Description{Side-by-side t-SNE plots for six classes without boundary calibration and with the proposed calibration. The calibrated representation separates several highlighted overlapping class pairs and produces tighter local clusters.}
   \label{fig:tsne}
\end{figure}

The t-SNE visualization in Fig.~\ref{fig:tsne} further demonstrates that, without boundary calibration, several high-risk class pairs exhibit clear bidirectional overlap, whereas our calibration separates these local conflict regions and yields cleaner class partitions. Together, these results show that prototype initialization dictates where a novel class enters the classifier space, while bidirectional calibration governs how surrounding decision regions evolve during adaptation.

\section{Conclusion}
In this paper, we study G-FSCIL, a challenging setting where the base session provides only weak priors for subsequent few-shot adaptation. We show that its key difficulty lies not only in data scarcity, but also in exploiting trustworthy synthetic knowledge without disrupting old--new decision boundaries. To address this, we propose a unified framework that combines trustworthy synthetic knowledge curation with boundary-stable incremental adaptation, enabling reliable support selection, robust novel-class initialization, and stable old--new separation. Extensive experiments on CIFAR100, CUB200, and ImageNet-R demonstrate consistent improvements over existing FSCIL methods, particularly under long-horizon incremental learning with weak base priors.

\begin{acks}
This work was supported by the National Natural Science Foundation of China (NSFC) No. 62406037, Guangdong Natural Science Funds for Distinguished Young Scholars (Grant 2023B1515020097), the National Research Foundation Singapore under the AI Singapore Programme (AISG Award No: AISG4-TC-2025-018-SGKR).
\end{acks}

\bibliographystyle{ACM-Reference-Format}
\balance
\bibliography{sample-base}

\clearpage
\nobalance
\appendix
\section*{Supplementary Material}
\suppressfloats[t]
\setcounter{figure}{0}
\setcounter{table}{0}
\renewcommand{\thefigure}{S\arabic{figure}}
\renewcommand{\thetable}{S\arabic{table}}
\noindent\textbf{Overview.}
This supplementary material provides additional experimental evidence and implementation details to complement the main paper. 
We first present detailed hyperparameter configurations and implementation settings to facilitate reproducibility. 
We then report additional quantitative results, including comparisons with state-of-the-art methods and comprehensive component ablations. 
Furthermore, we provide qualitative visualizations and hyperparameter sensitivity analyses to offer deeper insights into the behavior of the proposed method. 
\stitle{A. Experimental Setup}

\noindent\textbf{A.1. Hyperparameter Settings.}
Table~\ref{tab:supp_hp} summarizes the detail hyperparameter settings used on different datasets.

\begin{table}[H]
\centering
\scriptsize
\caption{Detailed hyperparameter settings on different datasets.}
\label{tab:supp_hp}
\setlength{\tabcolsep}{3.5pt}
\renewcommand{\arraystretch}{1.08}
\resizebox{\columnwidth}{!}{
\begin{tabular}{lccc}
\toprule
\multirow{2}{*}{Hyperparameter} & \multicolumn{3}{c}{Dataset} \\
\cmidrule(lr){2-4}
& CIFAR-100 & CUB-200 & ImageNet-R \\
\midrule
Backbone & ViT-B/16 & ViT-B/16 & ViT-B/16 \\
Prompt type & Deep VPT & Deep VPT & Deep VPT \\
Prompt length & 6 & 6 & 6 \\
Base epochs & 10 & 10 & 10 \\
Incremental epochs & 10 & 10 & 10 \\
Base learning rate & 0.01 & 0.05 & 0.02 \\
Incremental learning rate & 0.01 & 0.01 & 0.01 \\
Optimizer & SGD & SGD & SGD \\
Weight decay & 5e-4 & 5e-4 & 5e-4 \\
Base batch size & 5 & 7 & 10 \\
Incremental batch size & 5 & 5 & 5 \\
Diversity weight $\lambda_{\mathrm{div}}$ & 0.1 & 0.1 & 0.1 \\
Selection ratio $\rho_{\mathrm{sel}}$ & 0.01 & 0.001 & 0.01 \\
Mask regularization $\beta$ & 2 & 5.8 & 4 \\
Prototype fusion ratio $\rho_c$ & 0.1 & 0.1 & 0.1 \\
Forward weight $\lambda_{\mathrm{fwd}}$ & 0.5 & 0.5 & 0.5 \\
Reverse weight $\lambda_{\mathrm{rev}}$ & 0.01 & 0.01 & 0.01 \\
Calibration margin $\delta$ & 0.1 & 0.1 & 0.1 \\
\bottomrule
\end{tabular}
}
\vspace{-0.4em}
\end{table}

\noindent\textbf{A.2. Implementation Details.}
All experiments are implemented in PyTorch and conducted on a single NVIDIA RTX 4090 GPU. 
We follow the G-FSCIL protocol described in the main paper, where the model is first trained on the base session with a small number of base classes and then incrementally adapted to novel sessions. 
The same backbone architecture and pretrained initialization are used across all datasets to ensure fair comparison. 
Unless otherwise specified, the dataset-specific hyperparameters are listed in Table~\ref{tab:supp_hp}. 
We report AvgAcc, HAcc, and PD following prior FSCIL literature.


\stitle{B. Additional Results}

\noindent\textbf{B.1. Comparison with State-of-the-Art Methods on CUB200.}

\begin{table*}[t]
\centering
\caption{Comparison with state-of-the-art FSCIL methods on the CUB200 dataset.}
\label{tab:cub200}
\footnotesize
\setlength{\tabcolsep}{2pt}
\renewcommand{\arraystretch}{1.05}
\resizebox{\textwidth}{!}{
\begin{tabular}{lcccccccccccccccccccc@{\hspace{4pt}}c@{\hspace{6pt}}cc}
\toprule
\multirow{2}{*}{Method} & \multicolumn{20}{c}{Acc. in each session $\uparrow$} & \multirow{2}{*}{PD $\downarrow$} & \multirow{2}{*}{HAcc $\uparrow$} & \multirow{2}{*}{AvgAcc $\uparrow$} \\
\cmidrule(lr){2-21}
& 0 & 1 & 2 & 3 & 4 & 5 & 6 & 7 & 8 & 9 & 10 & 11 & 12 & 13 & 14 & 15 & 16 & 17 & 18 & 19 & & & \\
\midrule

Coda-Prompt~\cite{smith2023codaprompt}
& 95.0 & 53.9 & 21.9 & 18.2 & 13.9 & 7.5 & 8.2 & 7.8 & 6.8 & 7.1
& 6.0 & 5.5 & 4.5 & 4.0 & 3.4 & 3.8 & 2.7 & 3.6 & 4.4 & 3.7
& 91.3 & 0.0 & 14.9 \\

DualPrompt~\cite{wang2022dualprompt}
& 90.1 & 88.9 & 80.2 & 73.3 & 61.8 & 55.0 & 50.3 & 45.4 & 42.0 & 38.9
& 36.3 & 34.9 & 32.7 & 30.4 & 28.5 & 27.8 & 26.7 & 25.2 & 23.9 & 22.3
& 67.8 & 14.5 & 44.3 \\

L2P~\cite{wang2022l2p}
& \textbf{96.8} & 83.5 & 63.6 & 39.3 & 30.2 & 17.2 & 15.7 & 9.6 & 8.1 & 7.7
& 10.7 & 8.5 & 7.4 & 6.8 & 7.0 & 5.0 & 4.7 & 5.4 & 5.1 & 5.0
& 91.8 & 9.9 & 22.4 \\

DDGR~\cite{gao2023ddgr}
& 40.3 & 25.2 & 15.6 & 10.9 & 10.9 & 9.5 & 6.9 & 6.6 & 5.5 & 5.1
& 4.4 & 4.1 & 4.1 & 3.7 & 3.2 & 2.8 & 2.6 & 2.0 & 2.0 & 2.0
& 38.3 & 2.8 & 7.8 \\

\midrule

LGSP-Prompt~\cite{jiang2025lgsp}
& 88.1 & 85.4 & 74.9 & 72.2 & 68.0 & 58.2 & 29.0 & 12.9 & 8.0 & 5.1
& 5.2 & 2.1 & 2.0 & 2.7 & 2.8 & 1.4 & 1.7 & 0.8 & 1.6 & 1.2
& 86.8 & 0.0 & 26.1 \\

TEEN~\cite{wang2023few}
& 77.9 & 33.7 & 22.8 & 22.0 & 19.6 & 17.0 & 15.3 & 13.3 & 12.0 & 10.6
& 10.4 & 9.3 & 8.6 & 8.7 & 7.5 & 7.1 & 6.7 & 6.3 & 6.2 & 5.9
& 72.0 & 2.4 & 15.8 \\

SEC-Prompt~\cite{liu2025sec}
& 97.5 & 89.3 & 79.3 & 76.0 & 77.9 & 76.0 & 76.6 & 77.2 & 75.7 & 76.3
& 76.9 & 71.9 & 74.8 & 73.8 & 74.0 & 74.3 & 74.6 & 74.9 & 74.8 & 74.8
& 22.7 & 75.7 & 76.4 \\

ASP~\cite{liu2024few}
& 93.8 & 92.0 & 84.6 & 82.5 & 80.8 & 79.8 & 76.0 & 77.2 & 78.2 & 78.5
& 76.5 & 75.7 & 74.1 & 74.0 & 72.9 & 72.8 & 73.0 & 72.4 & 73.0 & 73.4
& 20.5 & 79.8 & 77.6 \\

\midrule

\textbf{Ours}
& 96.7 & \textbf{92.8} & \textbf{86.0} & \textbf{84.2} & \textbf{83.7} & \textbf{83.0} & \textbf{79.5} & \textbf{80.2} & \textbf{80.1} & \textbf{80.4}
& \textbf{78.9} & \textbf{78.4} & \textbf{76.7} & \textbf{77.3} & \textbf{76.0} & \textbf{75.4} & \textbf{75.7} & \textbf{74.0} & \textbf{75.0} & \textbf{75.7}
& \textbf{20.2} & \textbf{79.9} & \textbf{80.9} \\

\bottomrule
\end{tabular}
}
\end{table*}

Table~\ref{tab:cub200} reports detailed comparisons with representative FSCIL methods on CUB200. 
Our method achieves the best overall performance, reaching 80.9 AvgAcc, 79.9 HAcc, and 20.2 PD. 
Compared with the strongest prior baseline ASP, our method improves AvgAcc by 3.3 points while further reducing PD by 0.3. 
Compared with SEC-Prompt, it brings gains of 4.5 points in AvgAcc and 4.2 points in HAcc, together with a 2.5-point reduction in PD. 
Although the proposed method does not achieve the highest base-session accuracy, it consistently outperforms prior approaches from the first incremental session onward, indicating a more favorable balance between plasticity and long-term stability.

Overall, these results suggest that the proposed framework is particularly effective at preserving class discrimination as new classes arrive, and that its advantages become increasingly evident in more challenging incremental regimes where both reliable synthetic curation and stable prototype calibration are critical.


\noindent\textbf{B.2. Component Ablation Study.}

We further provide detailed ablations of the three main components in our framework, including \textbf{TSKC}, \textbf{Si-PI}, and \textbf{Bi-BC}. 
Here, TSKC denotes the base-stage trustworthy synthetic knowledge curation, which includes adaptive synthetic curation together with the transferable policy scorer; 
Si-PI denotes the synthetic-informed prototype initialization; and Bi-BC denotes the bidirectional boundary calibration during incremental adaptation. 
Tables~\ref{tab:ablation_cifar} and~\ref{tab:ablation_cub} summarize the corresponding results on CIFAR100 and CUB200, respectively.

On \textbf{CIFAR100}, TSKC already serves as the most effective single component, achieving 78.99 AvgAcc and 77.92 HAcc. 
This demonstrates that trustworthy synthetic knowledge curation in the base stage is essential for stabilizing the low-base regime. 
Adding Si-PI on top of TSKC further improves performance to 79.87 AvgAcc and 79.68 HAcc, while also achieving the best old-class accuracy (93.80), indicating improved representation stability and old-class preservation. 
In contrast, combining TSKC with Bi-BC leads to a larger gain in incremental performance, boosting AvgAcc to 86.36 and HAcc to 83.46, and achieving the best new-class accuracy (78.42). 
This suggests that once reliable synthetic knowledge has been established, boundary calibration becomes particularly effective in resolving old--new conflicts during incremental adaptation. 
Finally, integrating all three components yields the best overall result, reaching 86.89 AvgAcc, 84.23 HAcc, and 20.37 PD.

On \textbf{CUB200}, the ablation exhibits a more delicate pattern due to the fine-grained nature of the dataset. 
Configurations that introduce synthetic-assisted incremental adaptation without TSKC are highly unstable, leading to extremely low old-class accuracy and HAcc. 
This directly shows that, on fine-grained data, uncurated synthetic knowledge can be harmful and that trustworthy curation is a necessary prerequisite for subsequent incremental use. 
Once TSKC is enabled, performance improves dramatically: TSKC alone reaches 78.36 AvgAcc and 81.80 HAcc, already establishing a strong baseline. 
Adding Si-PI on top of TSKC further improves HAcc to 82.02 and old-class accuracy to 90.53, indicating that synthetic-informed prototype initialization is particularly beneficial for stabilizing fine-grained class insertion and preserving previously learned categories. 
By contrast, combining TSKC with Bi-BC yields the best new-class accuracy (75.84) and the lowest PD (18.76), while also improving AvgAcc to 80.25, suggesting that boundary calibration is more effective in enhancing plasticity and reducing performance degradation. 
When all three components are jointly used, the model achieves the best AvgAcc of 80.44 while maintaining competitive HAcc, old-class accuracy, and new-class accuracy, demonstrating the strongest overall end-to-end configuration.

Overall, these results are consistent with the design of our method. 
TSKC provides a trustworthy synthetic foundation, Si-PI mainly stabilizes prototype initialization and old-class retention, and Bi-BC mainly improves old--new boundary compatibility during incremental adaptation. 
Their combination yields the most robust performance across datasets, while the relative contribution of each module varies with dataset difficulty and granularity.

\renewcommand{\arraystretch}{0.9}
\begin{table}[t]
\centering
\small
\caption{Ablation study on CIFAR100.}
\label{tab:ablation_cifar}
\begin{tabular}{c@{\hspace{3pt}}c@{\hspace{3pt}}c|ccccc}
\toprule
\multicolumn{3}{c|}{\textbf{Modules}} & \multicolumn{5}{c}{\textbf{Metrics}} \\
\cmidrule(lr){1-3} \cmidrule(lr){4-8}
TSKC & Si-PI & Bi-BC & AvgAcc $\uparrow$ & HAcc $\uparrow$ & Old $\uparrow$ & New $\uparrow$ & PD $\downarrow$ \\
\midrule
 & \checkmark & \checkmark & 76.49 & 74.41 & 83.40 & 67.17 & 31.44\\
 &  & \checkmark & 78.05 & 73.31 & 79.60 & 67.94 & 30.67 \\
 & \checkmark &  & 78.35 & 76.90 & 92.20 & 65.96 & 32.12 \\
\checkmark &  &  & 78.99 & 77.92 & 93.00 & 67.05 & 30.65 \\
\checkmark & \checkmark &  & 79.87 & 79.68 & \textbf{93.80} & 69.26 & 28.91 \\
\checkmark &  & \checkmark & 86.36 & 83.46 & 89.20 & \textbf{78.42} & 20.74 \\
\checkmark & \checkmark & \checkmark & \textbf{86.89} & \textbf{84.23} & 91.00 & 78.39 & \textbf{20.37} \\
\bottomrule
\end{tabular}
\end{table}

\begin{table}[t]
\centering
\caption{Ablation study on CUB200.}
\small
\label{tab:ablation_cub}
\begin{tabular}{c@{\hspace{3pt}}c@{\hspace{3pt}}c|ccccc}
\toprule
\multicolumn{3}{c|}{\textbf{Modules}} & \multicolumn{5}{c}{\textbf{Metrics}} \\
\cmidrule(lr){1-3} \cmidrule(lr){4-8}
TSKC & Si-PI & Bi-BC & AvgAcc $\uparrow$ & HAcc $\uparrow$ & Old $\uparrow$ & New $\uparrow$ & PD $\downarrow$ \\
\midrule
 & \checkmark & \checkmark & 64.61 & 9.16 & 4.94 & 63.62 & 27.65 \\
 &  & \checkmark & 65.21 & 7.00 & 3.70 & 64.91 & 26.45 \\
 & \checkmark &  & 68.81 & 70.85 & 80.25 & 63.43 & 24.77 \\
\checkmark &  &  & 78.36 & 81.80 & 90.12 & 74.89 & 19.12 \\
\checkmark & \checkmark &  & 78.68 & \textbf{82.02} & \textbf{90.53} & 74.97 & 19.86 \\
\checkmark &  & \checkmark & 80.25 & 78.18 & 80.66 & \textbf{75.84} & \textbf{18.76} \\
\checkmark & \checkmark & \checkmark & \textbf{80.44} & 79.94 & 85.19 & 75.31 & 20.15\\
\bottomrule
\end{tabular}
\end{table}
\renewcommand{\arraystretch}{1.05}

\stitle{C. Analysis and Visualization}

\noindent\textbf{C.1. Qualitative Visualization of Synthetic Sample Curation.}

\begin{figure}[t]
    \centering
    \includegraphics[width=0.9\columnwidth]{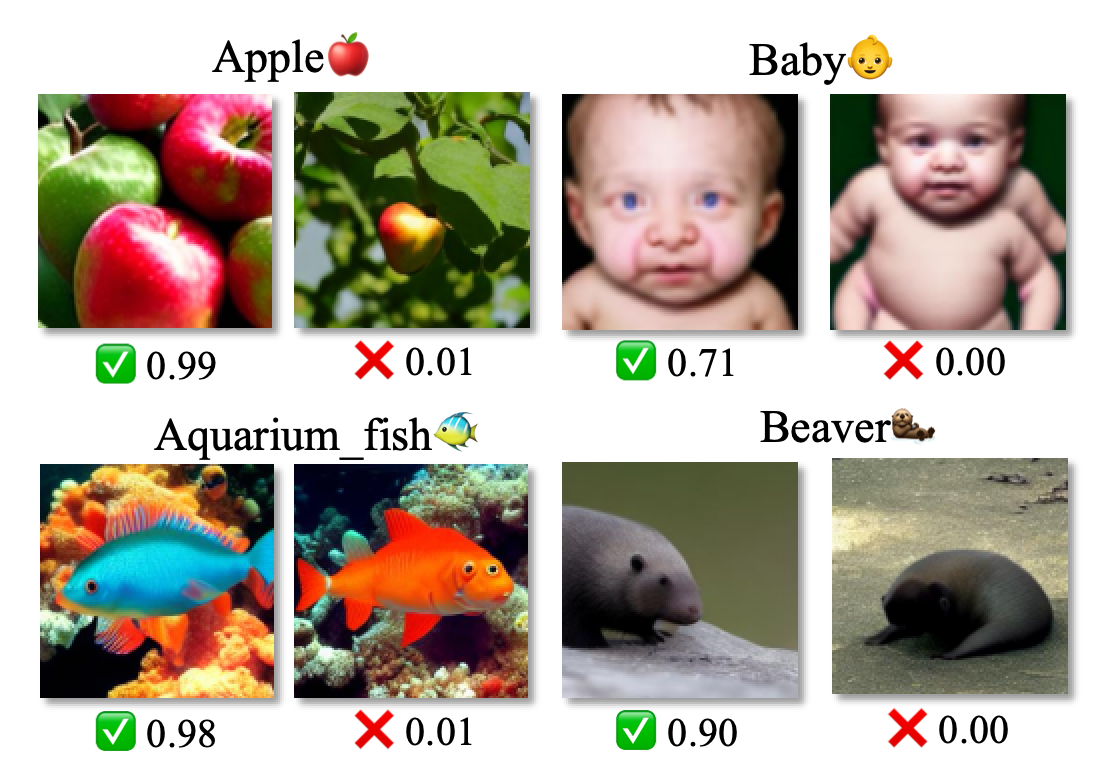}
    \caption{Qualitative visualization of synthetic sample curation. For each class, we show a representative retained sample (left, \checkmark) and a rejected sample (right, \texttimes), together with their reliability scores.}
    \label{fig:chosen}
\end{figure}

Figure~\ref{fig:chosen} visualizes representative examples of the synthetic-sample curation process. 
For each class, the proposed reliability scorer assigns substantially higher confidence to semantically correct and visually representative synthetic samples, while suppressing low-quality or ambiguous generations. 
The retained samples exhibit clear category identity and align well with the target semantics, whereas the rejected ones typically suffer from semantic mismatch, weak foreground structure, or poor visual fidelity. 
These observations demonstrate that the proposed curation strategy effectively filters unreliable synthetic samples before they participate in prototype construction, thereby reducing the risk of noisy synthetic replay.

\noindent\textbf{C.2. Visualization of Class-Discriminative Regions.}

\begin{figure}[b]
    \centering
    \includegraphics[width=0.9\columnwidth]{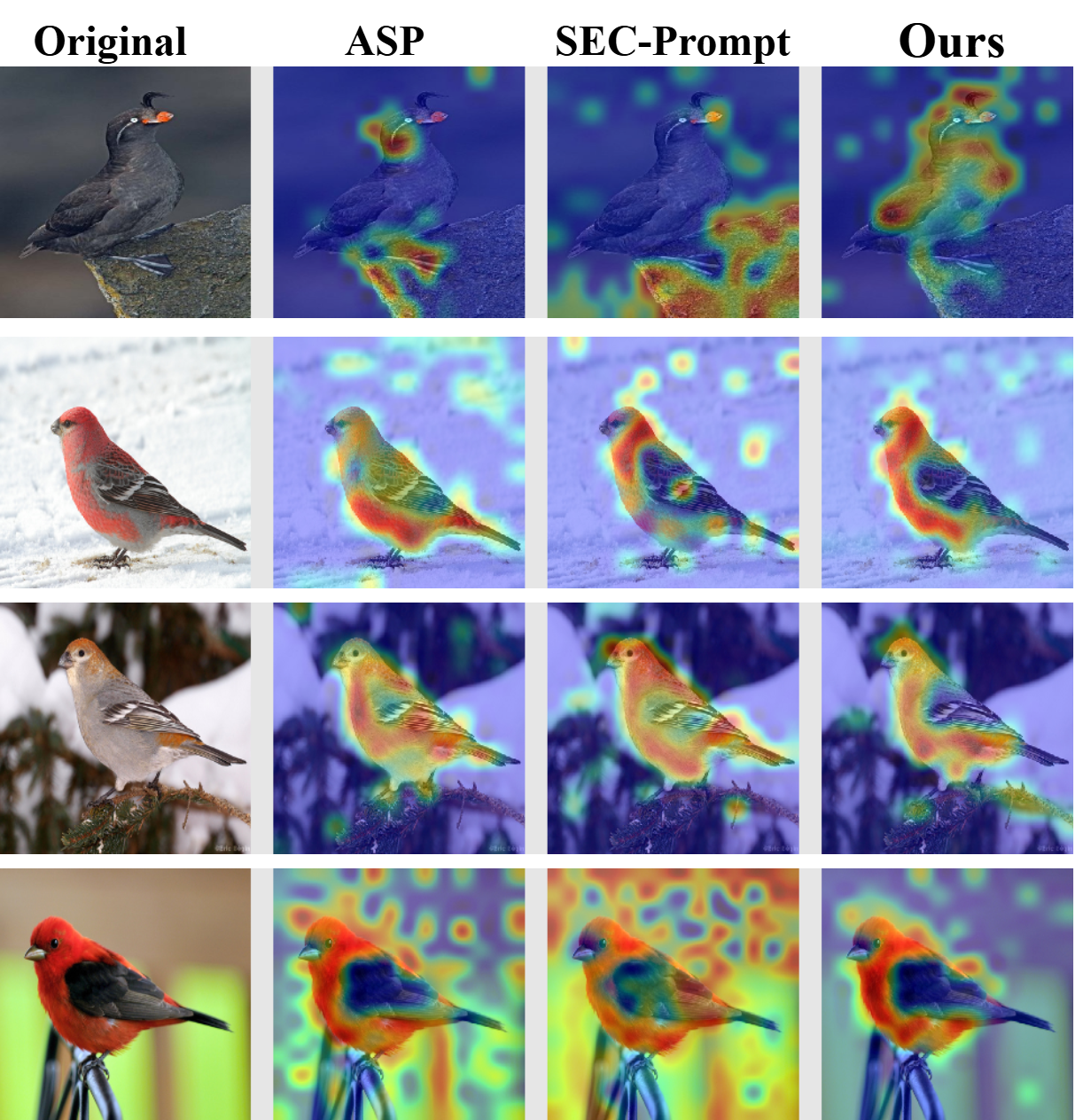}
    \caption{Visualization of class-discriminative activation maps.}
    \label{fig:heatmap}
\end{figure}

Figure~\ref{fig:heatmap} compares the discriminative regions highlighted by ASP, SEC-Prompt, and our method. 
Compared with prior methods, our model places more concentrated attention on the object body and other class-discriminative local cues, while suppressing irrelevant background responses. 
In challenging fine-grained bird examples, ASP and SEC-Prompt often exhibit distracted activations on surrounding context or overly dispersed responses, whereas our method yields more compact and object-centered localization. 
This observation suggests that the proposed framework not only improves feature discrimination, but also encourages the model to rely on category-relevant visual evidence, which is particularly important under low-base and few-shot incremental settings.

\noindent\textbf{C.3. Hyperparameter Sensitivity Analysis.}

\begin{figure}[b]
   \centering
   \includegraphics[width=\linewidth]{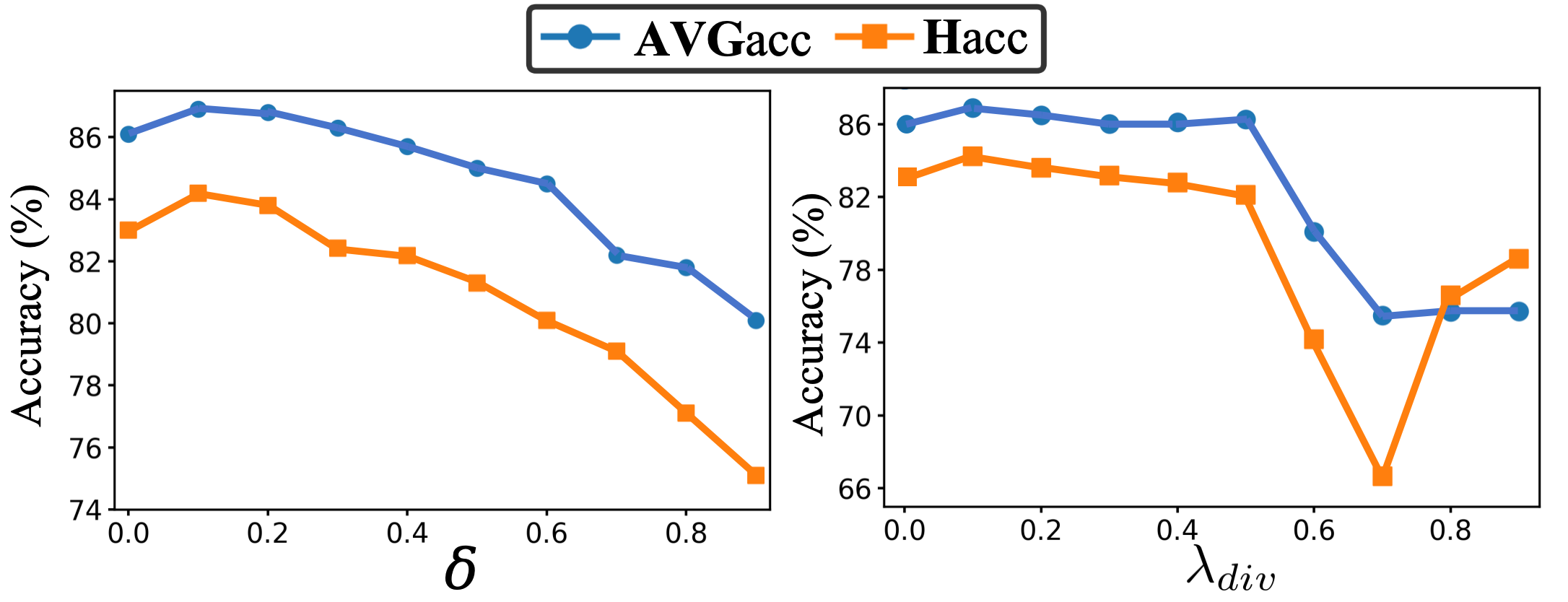}
   \caption{Sensitivity analysis of $\delta$ and $\lambda_{\mathrm{div}}$.}
   \label{fig:hparam_sensitivity}
\end{figure}

We further study the sensitivity of the proposed method to the calibration margin $\delta$ and the diversity weight $\lambda_{\mathrm{div}}$, as shown in Fig.~\ref{fig:hparam_sensitivity}.

For the calibration margin $\delta$, both AvgAcc and HAcc first improve and then gradually decline as $\delta$ increases. 
The best overall performance is achieved around $\delta=0.1$, where both metrics reach their peak or near-peak values. 
When $\delta$ becomes larger, the performance drops steadily, and the degradation becomes more evident after $\delta \geq 0.6$. 
This suggests that a moderate margin is sufficient to encourage effective boundary calibration, whereas an excessively large margin tends to over-separate classes and harm the stability--plasticity trade-off. 
We therefore set $\delta=0.1$ in all experiments.

For the diversity weight $\lambda_{\mathrm{div}}$, the method remains stable under relatively small values and achieves the best overall trade-off around $\lambda_{\mathrm{div}}=0.1$. 
As $\lambda_{\mathrm{div}}$ further increases, both AvgAcc and HAcc gradually deteriorate, followed by a sharp drop when $\lambda_{\mathrm{div}} \geq 0.6$. 
This behavior indicates that moderate diversity regularization helps prevent redundant synthetic selection, while overly strong regularization may disrupt semantic consistency and weaken the reliability of selected synthetic samples. 
Based on this observation, we use $\lambda_{\mathrm{div}}=0.1$ as the default setting.

\end{document}